\documentclass[superscriptaddress]{revtex4-2}

\usepackage{textcomp}

\usepackage[utf8]{inputenc} 
\usepackage[T1]{fontenc}    
\usepackage{hyperref}       
\usepackage{url}            
\usepackage{booktabs}       
\usepackage{amsfonts}       
\usepackage{nicefrac}       
\usepackage{enumitem}
\usepackage{microtype}      
\usepackage{xcolor} 
\usepackage{graphicx}
\usepackage{amsmath,amssymb,amsthm,mathrsfs, bm}
\usepackage{mathtools}
\usepackage{latexsym,amscd,amsbsy,dsfont,amsfonts}
\usepackage{physics}
\usepackage{bbm}
\usepackage[sort&compress]{natbib}

\newcommand\samples{n}
\newcommand\sdim{p}

\newcommand\reg{\lambda}
\newcommand\eig{\eta}
\newcommand\teacher{\theta^\star}
\newcommand\hilbert{\mathcal{H}}
\newcommand\feature{\psi}
\newcommand\featuredim{p}
\newcommand\regdecay{\ell}
\newcommand\featindex{k}
\newcommand\tot{\mathrm{tot}}
\newcommand\measure{\nu}
\newtheorem{theorem}{Theorem}

\begin{document}
\title{Generalization Error Rates in Kernel 
Regression:\\ The Crossover from the Noiseless to Noisy Regime}

%

\author{
  Hugo Cui 
}
\affiliation{SPOC, EPFL, Switzerland}
\author{
  Bruno Loureiro
}
\affiliation{IDePHICS lab., EPFL, Switzerland}
\author{
  Florent Krzakala
}
\affiliation{IDePHICS lab., EPFL, Switzerland}
\author{
  Lenka Zdeborov\'a
}
\affiliation{SPOC, EPFL, Switzerland}

\begin{abstract}
  In this manuscript we consider Kernel Ridge Regression (KRR) under the Gaussian design. Exponents for the decay of the excess generalization error of KRR have been reported in various works under the assumption of power-law decay of eigenvalues of the features co-variance. These decays were, however, provided for sizeably different setups, namely in the noiseless case with constant regularization and in the noisy optimally regularized case. Intermediary settings have been left substantially uncharted. In this work, we unify and extend this line of work, providing characterization of all regimes and excess error decay rates that can be observed in terms of the interplay of noise and regularization. In particular, we show the existence of a transition in the noisy setting between the noiseless exponents to its noisy values as the sample complexity is increased. Finally, we illustrate how this crossover can also be observed on real data sets.
\end{abstract}

\maketitle

\section{Introduction}

Kernel methods are among the most popular models in machine learning. Despite their relative simplicity, they define a powerful framework in which non-linear features can be exploited without leaving the realm of convex optimisation. Kernel methods in machine learning have a long and rich literature dating back to the 60s \cite{Nadaraja64, Watson64}, but have recently made it back to the spotlight as a proxy for studying neural networks in different regimes, e.g. the infinite width limit \cite{Neal1996, Williams1996,jacot2018neural, Lee2018} and the lazy regime of training \cite{Chizat2019}. Despite being defined in terms of a non-parametric optimisation problem, kernel methods can be mathematically understood as a standard parametric linear problem in a (possibly infinite) Hilbert space spanned by the kernel eigenvectors (a.k.a \emph{features}). This dual picture fully characterizes the asymptotic performance of kernels in terms of a trade-off between two key quantities: the relative decay of the eigenvalues of the kernel (a.k.a. its \emph{capacity}) and the coefficients of the target function when expressed in feature space (a.k.a. the \emph{source}). Indeed, a sizeable body of work has been devoted to understanding the decay rates of the excess error as a function of these two relative decays, and investigated whether these rates are attained by algorithms such as stochastic gradient descent \cite{pillaud2018statistical, Berthier2020TightNC}.

Rigorous optimal rates for the excess generalization error in kernel ridge regression are well-known since the seminal works of \cite{Caponnetto2005FastRF,steinwart2009optimal}. However, recent interesting works \cite{spigler2019asymptotic, bordelon2020} surprisingly reported very different - and actually better - rates supported by numerical evidences. These papers appeared to either not comment on this discrepancy \cite{bordelon2020}, or to attribute this apparent contradiction to a difference between typical and worse-case analysis~\cite{spigler2019asymptotic}. As we shall see, the key difference between these works stems instead from the fact that most of classical works considered {\it noisy} data and fine-tuned regularization, while \cite{spigler2019asymptotic, bordelon2020} focused on noiseless data sets. This observation raises a number of questions: is there a connection between both sets of exponents? Are Gaussian design exponents actually different from worst-case ones?  What about intermediary setups (for instance noisy labels with generic regularization, noiseless labels with varying regularization) and regimes (intermediary sample complexities)? How does infinitesimal noise differ from no noise at all?

\paragraph{Main contributions ---} In this manuscript, we answer all the above questions, and redeem the apparent contradiction by reconsidering the Gaussian design analysis. We provide a unifying picture of the decay rates for the excess generalization error, along a more exhaustive characterization of the regimes in which each is observed, evidencing the interplay of the role of regularization, noise and sample complexity. We show in particular that typical-case analysis with a Gaussian design is actually in perfect agreement with the statistical learning worst-case data-agnostic approach. We also show how the optimal excess error decay can transition from the recently reported noiseless value to its well known noisy value as the number of samples is increased. We illustrate this crossover from the \textit{noiseless} regime to the \textit{noisy} regime also in a variety of KRR experiments on real data. 

\paragraph{Related work ---} The analysis for kernel methods and ridge regression is a classical topic in statistical learning theory~\cite{Caponnetto2005FastRF,caponnetto2007optimal,steinwart2009optimal,fischer2017sobolev,Lin2018OptimalRF,Bartlett2020BenignOI,Lin2018OptimalRF}. In this classical setting, decay exponents for optimally regularized {\it noisy} linear regression on features with power-law co-variance spectrum have been provided. Interestingly, it has been shown that such optimal rates can be obtained in practice by stochastic gradient descent, without explicit regularization, with single-pass ~\cite{polyak1992acceleration,nemirovskij1983problem} or multi-pass~\cite{pillaud2018statistical},  as well as by randomized algorithms \cite{Jun2019KernelTR}. Closed-form bounds for the prediction error have been provided in a number of worst-case analyses \cite{Jun2019KernelTR,Lin2018OptimalRF}. We show how the decay rates given in the present paper can also be alternatively deduced therefrom in Appendix \ref{appendix:related}.

The recent line of work on the noiseless setting includes contributions from statistical learning theory \cite{Berthier2020TightNC, Varre2021LastIC} and statistical physics \cite{spigler2019asymptotic, bordelon2020}. This much more recent second line of work proved decay rates for a given, constant regularization. An example of noise-induced crossover is furthermore mentioned in \cite{Berthier2020TightNC}. The interplay between noisy and noiseless regimes has also been investigated in the related Gaussian Process literature \cite{Kanagawa2018}.

The study of ridge regression with Gaussian design is also a classical topic. Ref.~\cite{dicker2016ridge} considered a model in which the covariates are isotropic Gaussian in ${\mathbb R}^p$, and computed the exact asymptotic generalization error in the high-dimensional asymptotic regime $p, n \to \infty$ with dimension-to-sample-complexity ratio $p/n$ fixed. This result was generalised to arbitrary co-variances \cite{Hsu2012, dobriban2018high} using fundamental results from random matrix theory \cite{ledoit2011eigenvectors}. Non-asymptotic rates of convergence for a related problems were given in \cite{Loureiro2021CapturingTL}. Previous results also existed in the statistical physics literature, e.g. \cite{dietrich1999statistical,opper1996statistical, Opper2001, kabashima2008inference}. 
Gaussian models for regression have seen a surge of popularity recently, and have been used in particular to study over-parametrization and the double-descent phenomenon, e.g. in \cite{advani2020high,belkin2020two,hastie2019surprises,mei2019generalization,  gerace2020generalisation,ghorbani2019limitations,kobak2020optimal,wu2020optimal,Bartlett2020BenignOI,richards2021asymptotics,liao2020random,jacot2020kernel,ghorbani2020neural, liu2020kernel}. 


\section{Setting}

\label{section:setting}
Consider a data set $\mathcal{D}=\{x^\mu,y^\mu\}_{\mu=1}^n$ with $n$ independent samples from a probability measure $\measure$ on $\mathcal{X}\times \mathcal{Y}$, where $\mathcal{X}\subset \mathbb{R}^{d}$ is the input and $\mathcal{Y} \subset\mathbb{R}$ the response space. Let $K$ be a kernel and $\hilbert$ denote its associated reproducing kernel Hilbert space (RKHS). \emph{Kernel ridge regression} (KRR) corresponds to the following non-parametric minimisation problem: 
\begin{equation}
    \underset{f\in\hilbert}{\min}~ \frac{1}{n}\sum\limits_{\mu=1}^n (f(x^\mu)-y^\mu)^2+\reg ||f||^2_\hilbert{}.
\label{eq:KRR_risk:hilbert}
\end{equation}
\noindent where $||\cdot||_{\hilbert}$ is the norm associated with the scalar product in $\hilbert$, and $\reg \geq 0$ is the regularisation. The convenience of KRR is that it admits a dual representation in terms of a standard parametric problem. Indeed, the kernel $K$ can be diagonalized in an orthonormal basis $\{\phi_{\featindex}\}_{\featindex=1}^{\infty}$ of $L^2(\mathcal{X})$:
\begin{align}
    \int_{\mathcal{X}}\measure_{x}(\dd x') K(x, x') \phi_{\featindex}(x') = \eig_{\featindex} \phi_{\featindex}(x)
\end{align}
\noindent where $\{\eig_{\featindex}\}_{\featindex=1}^{\infty}$ are the corresponding (non-negative) kernel eigenvalues and $\measure_{x}$ is the marginal distribution over $\mathcal{X}$. Note that the kernel $\{\phi_{\featindex}\}_{\featindex=1}^{\infty}$ eigenvectors form an orthonormal basis of $L^2(\mathcal{X})$. It is convenient to define the re-scaled basis of \emph{kernel features} $\psi_{\featindex}(x) = \sqrt{\eig_{\featindex}} \phi_{\featindex}(x)$ and to work in matrix notation in feature space: define $\phi(x) \equiv \{\phi_{\featindex}(x)\}_{\featindex=1}^{p}$ (with $p$ possibly infinite)
\begin{align}
\label{eq:def_feature_map}
\feature{}(x)=\Sigma^{\frac{1}{2}}\phi(x)\, && \mathbb{E}_{x\sim \measure_{x}}\left[ \phi(x)\phi(x)^{\top}\right]=\mathbbm{1}_\featuredim{}\, , && \mathbb{E}_{x'\sim\measure_{x}}\left[K(x,x')\phi(x')\right]=\Sigma \phi(x)\, ,
\end{align}
\noindent where $\Sigma \equiv \mathbb{E}_{x\sim\measure_{x}}\left[\feature(x)\feature(x)^{\top}\right] = \mathrm{diag}(\eig_1,\eig_2,...,\eig_{\featuredim})$ is the features co-variance (a diagonal operator in feature space). In this notation, the RKHS $\hilbert$ can be formally written as $\hilbert=\{f=\feature{}^{\top}\theta: \theta\in\mathbb{R}^{\featuredim}, \quad ||\theta||_2 < \infty\}$, i.e. the space of functions for which the coefficients in the feature basis are square summable. With this notation, we can rewrite eq.~\eqref{eq:KRR_risk} in feature space as a standard parametric problem for the following empirical risk:
\begin{equation}
    \hat{\mathcal{R}}_{n}(w) = \frac{1}{n}\sum\limits_{\mu=1}^n \left(w^{\top}\feature(x^\mu) - y^\mu\right)^2+\reg~ w^{\top}w.
\label{eq:KRR_risk}
\end{equation}
Our main results concern the typical averaged performance of the KRR estimator, as measured by the typical prediction (out-of-sample) error 
\begin{equation}
    \epsilon_g = \mathbb{E}_{\mathcal{D}}\mathbb{E}_{(x,y)\sim\measure}(\hat{f}(x)-y)^2\, ,
\label{eq:excess}
\end{equation}
\noindent where the first average is over the data $\mathcal{D} = \{x^{\mu}, y^{\mu}\}$ and the second over a fresh sample $(x,y)\sim \measure$. 


In what follows we assume the labels $y^{\mu}\in\mathcal{Y}$ were generated, up to an independent additive Gaussian noise with variance $\sigma^2$, by a target function $f^{\star}$ (not necessarily belonging to $\hilbert$):
\begin{equation}
    y^\mu \overset{d}{=} f^{\star}(x^\mu)+\sigma \mathcal{N}(0,1),
    \label{eq:def:teacher}
\end{equation}
\noindent and we denote by $\teacher$ the coefficients of the target function in the features basis $f^{\star}(x) = \feature(x)^{\top}\teacher$. As we will characterize below, whether the target function $f^{\star}$ belongs or not to $\hilbert$ depends on the relative decay coefficients $\teacher$ with respect to the eigenvalues of the kernel. We often refer to $\teacher$ as the \emph{teacher}.  While the present results and discussion are provided for additive gaussian noise for simplicity, our method are not restricted to this particular noise, and a more complete extension of the results for other noise settings is left for future work.

We are then interested in the evolution of the \emph{excess error} $\epsilon_g-\sigma^2$ as the number of samples $n$ is increased.

\paragraph{Capacity and source coefficients ---} Motivated by the discussion above, we focus on ridge regression in an infinite dimensional ($\featuredim \to \infty$) space $\hilbert{}$ with Gaussian design $u^\mu \overset{\mathrm{def}}{=}\feature{}(x^\mu)\overset{d}{=}\mathcal{N}(0,\Sigma)$ with (without loss of generality) diagonal co-variance $\Sigma=\mathrm{diag}(\eig_1,\eig_2,...)$. We expect however the results of this manuscript to be universal for a large class of distribution beyond the Gaussian one. In particular, we anticipate the gaussianity assumption should be amenable to being relaxed to sub-gaussians \cite{Tsigler2020BenignOI} or even any concentrated distribution \cite{Talagrand1994ConcentrationOM,Louart2017ARM}.

Following the statistical learning terminology, we introduce two parameters $\alpha>1, r\ge0$, herefrom referred to as the \textit{capacity} and \textit{source} conditions \cite{caponnetto2007optimal}, to parametrize the difficulty of the target function and the learning capacity of the kernel
\begin{align}
    \mathrm{tr}~\Sigma^{\frac{1}{\alpha}}<\infty,&& ||\Sigma^{\frac{1}{2}-r}\theta^\star||_{\mathcal{H}} < \infty.
\label{eq:source_capa_conditions}
\end{align}
As in \cite{dobriban2018high,spigler2019asymptotic, bordelon2020, Berthier2020TightNC}, we consider the particular case where both the spectrum of $\Sigma$ and the teacher components $\teacher_\featindex{}$ have exactly a power-law form satisfying the limiting source/capacity conditions \eqref{eq:source_capa_conditions}:
\begin{align}
    \eig_\featindex{}= \featindex{}^{-\alpha}\, ,&&\teacher_\featindex{}=\featindex{}^{-\frac{1+\alpha (2r-1)}{2}}\, .
 \label{eq:model_def}
\end{align}

The power law ansatz \eqref{eq:model_def} is empirically observed to be a rather good approximation for some real simple datasets and kernels, see Fig.~\ref{fig:Measure_params} in Appendix \ref{appendix:real_data}. The parameters $\alpha, r$ introduced in \eqref{eq:model_def} control the complexity of the data the teacher respectively. A large $\alpha$ can be loosely seen as characterizing a effectively low dimensional  (and therefore easy to fit) data distribution. By the same token, a large $r$ signals a good alignment of the teacher with the important directions in the data covariance, and therefore an a priori simpler learning task.

The regularization $\reg$ is allowed to vary with $n$ according to a power-law $\reg=n^{-\regdecay}$. This very general form allows us to encompass both the zero regularization case (corresponding to $\regdecay=\infty$) and the case where $\reg=\reg^\star$ is optimized, with some optimal decay rate $\regdecay^\star$. Note that this power law form implies that $\lambda$ is assumed positive. While this is indeed the assumption of \cite{Caponnetto2005FastRF,caponnetto2007optimal} with which we intend to make contact, \cite{wu2020optimal} have shown that the optimal $\lambda$ may in some settings be negative. Some numerical experiments suggest that removing the positivity constraint on $\lambda$ while optimizing does not affect the results presented in this manuscript. A more detailed investigation is left to future work.

\section{Main results}

\label{section:phase_diagram}
\begin{figure}[b]
    \includegraphics[scale=0.7]{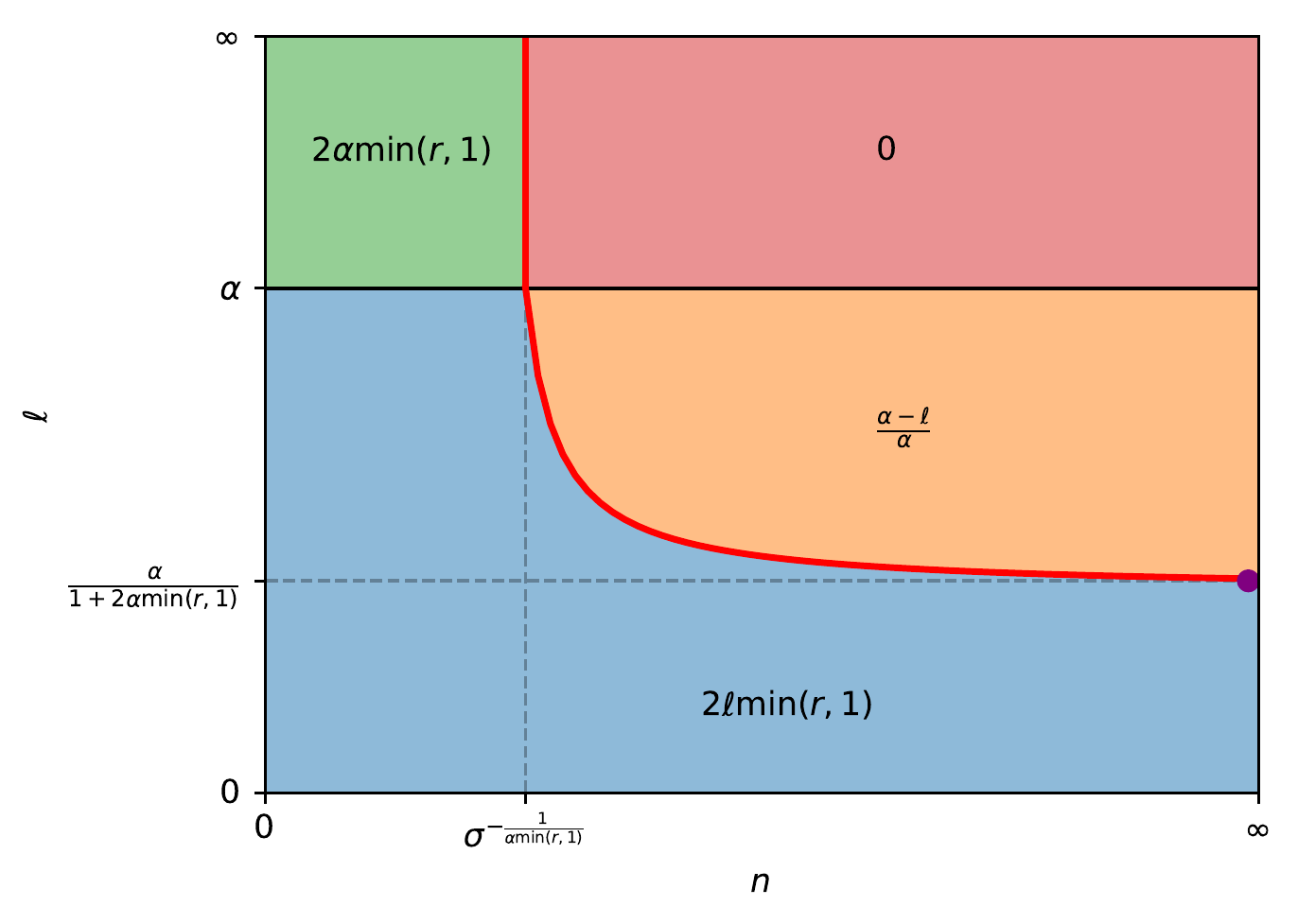}
    \caption{ \label{fig:phase_diagram}Different decays for the excess generalization error $\epsilon_g-\sigma^2$ for different values of $n$ and different decays $\regdecay$ of the regularization $\reg\sim n^{-\regdecay}$, at given noise variance $\sigma$. The red solid line represents the noise-induced crossover line, separating the effectively noiseless regime (green and blue) on its left from the effectively noisy regime (red and orange) on its right. Any KRR experiment at fixed regularization decay $\regdecay$ (corresponding to drawing a horizontal line at ordinate $\regdecay$) crosses the crossover line if $\regdecay>\alpha/(1+2\alpha\mathrm{min}(r,1))$. The corresponding learning curve will accordingly exhibit a crossover from a fast decay (noiseless regime) to a slow decay (noisy regime).}
\end{figure}

\begin{figure}[t]
    \centering
    \includegraphics[scale=0.5]{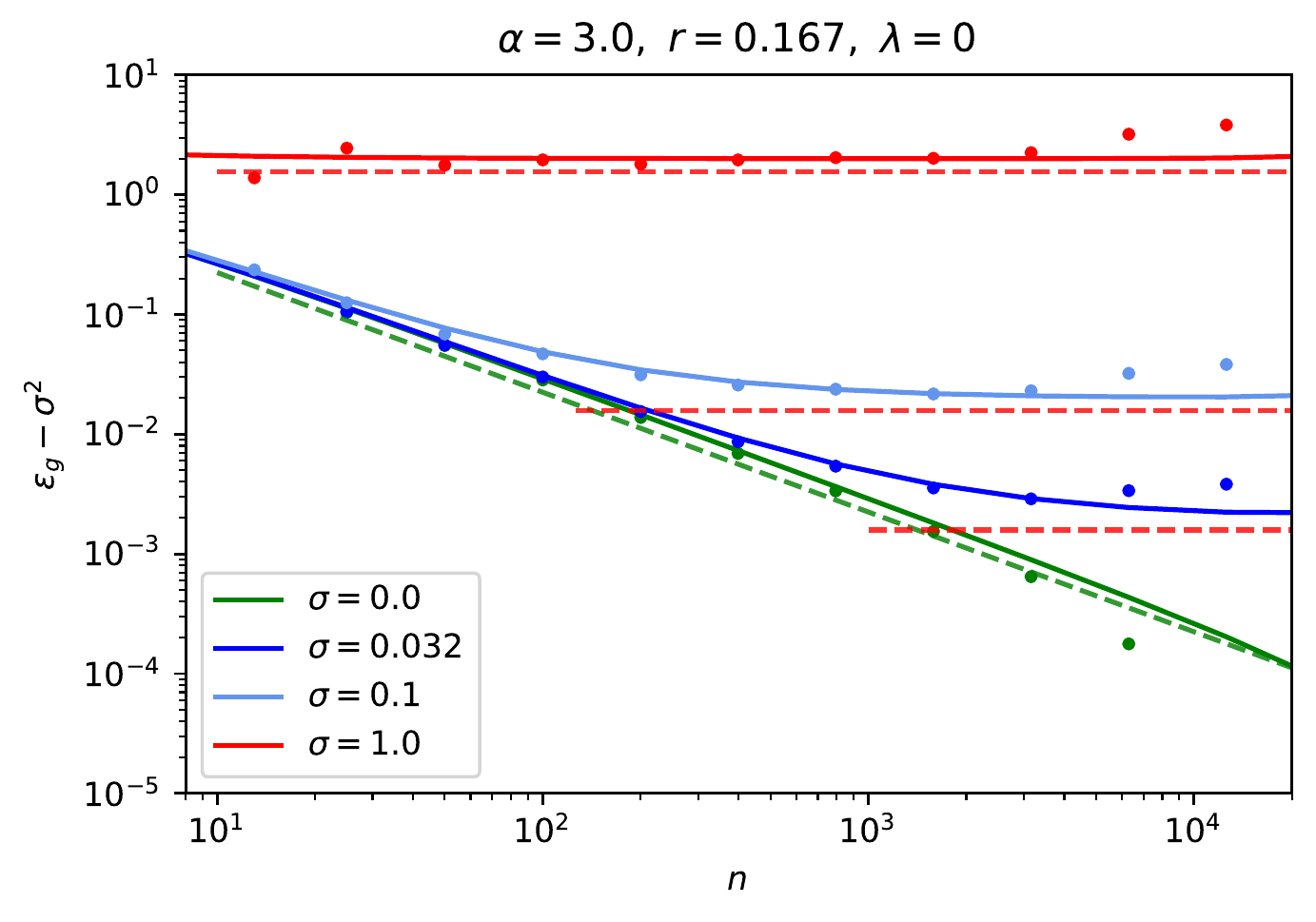}
    \includegraphics[scale=0.5]{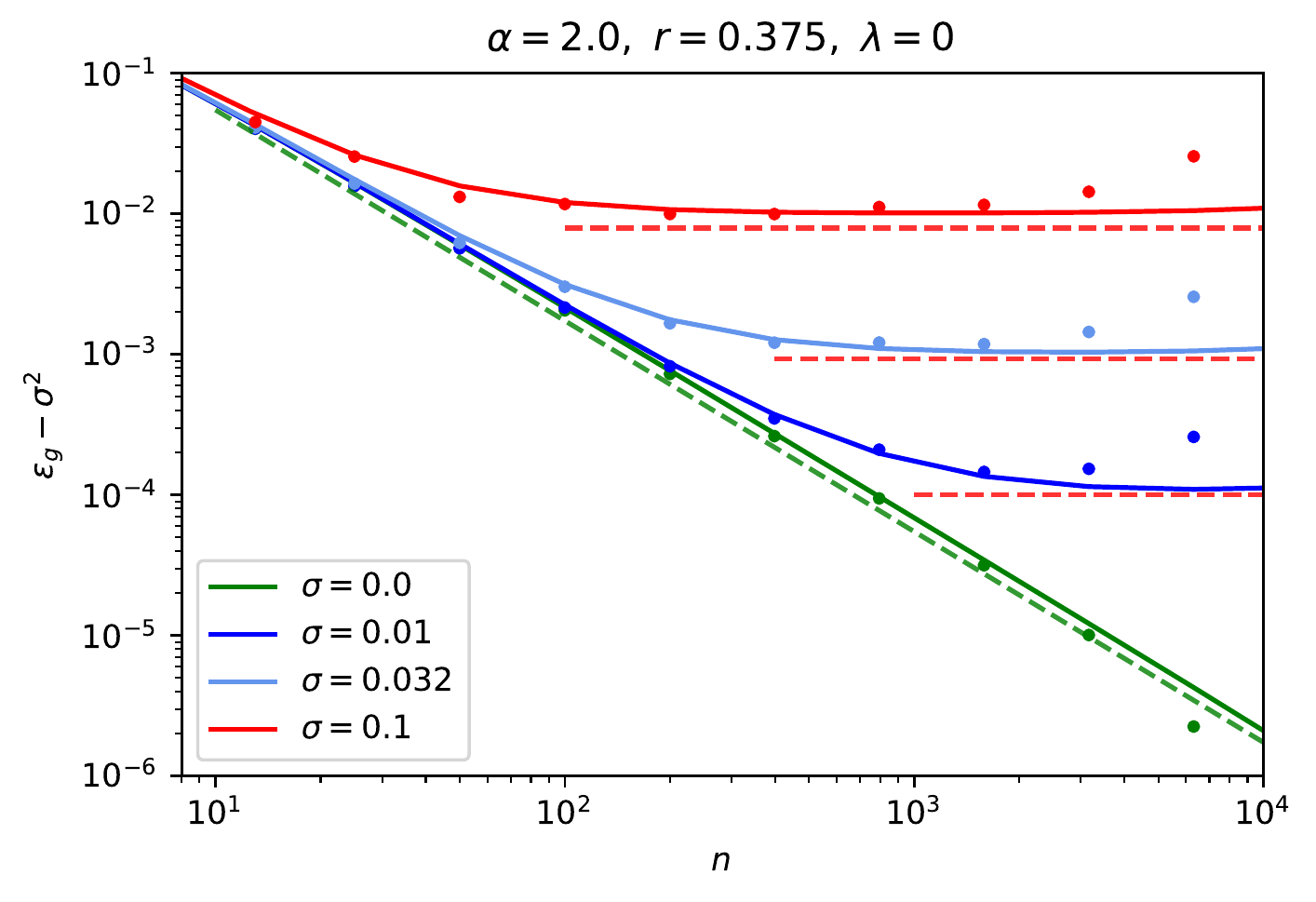}
    \caption{Kernel ridge regression on synthetic data sets with capacity $\alpha$ and source coefficient~$r$ i.e. the idealized gaussian setting \eqref{eq:model_def}}, with no regularization $\reg=0$. Solid lines correspond to the theoretical prediction of eq.~(\ref{eq:SP_kernel}) using the \texttt{GCM} package associated with \cite{Loureiro2021CapturingTL}. 
    Points are simulations conducted using the python \texttt{scikit-learn KernelRidge} package \cite{scikit-learn}, where the feature space dimension has been cut off to $\featuredim=10^4$ for the simulations, and to $10^5$ for the theoretical curves. Dashed lines represent the slopes predicted by eq.~\eqref{eq:green_red_decay}, with the color (red and green) in correspondence to the regime from Fig.~\ref{fig:phase_diagram}.
    \label{fig:Artificial_noreg}

\end{figure}

\begin{figure}[ht]
    \centering
    \includegraphics[scale=0.5]{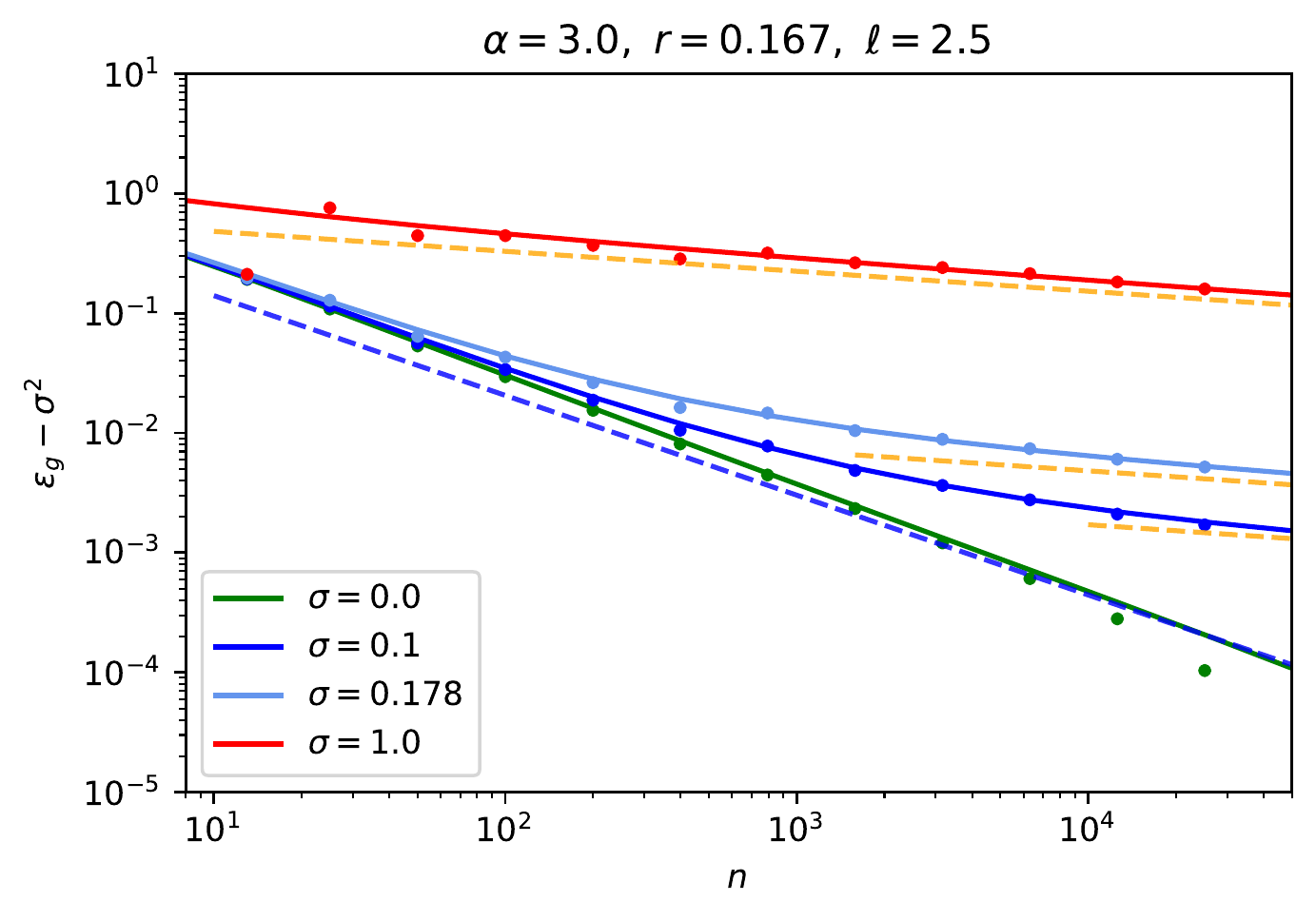}
    \includegraphics[scale=0.5]{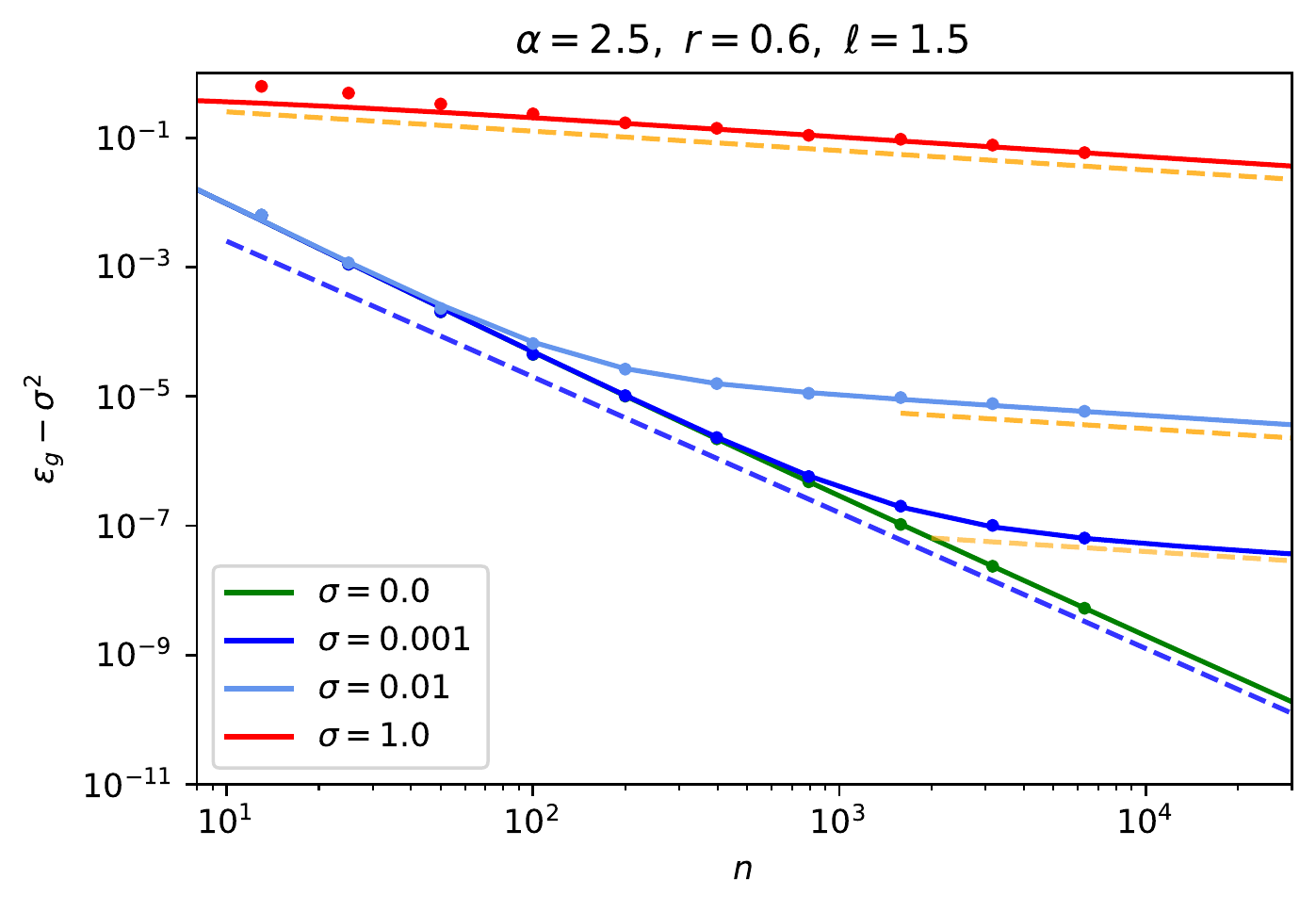}
    \caption{Kernel ridge regression on synthetic data sets with capacity $\alpha$ and source coefficient~$r$, with regularization $\reg=n^{-\regdecay}$. Solid lines correspond to the theoretical prediction of eq.~(\ref{eq:SP_kernel}) using the \texttt{GCM} package associated with \cite{Loureiro2021CapturingTL}. 
    Points are simulations conducted using the python \texttt{scikit-learn KernelRidge} package \cite{scikit-learn}, where the feature space dimension has been cut off to $\featuredim=10^4$ for the simulations, and to $10^5$ for the theoretical curves. Dashed lines represent the slopes predicted by eq.~\eqref{eq:blueorange_decay}, with the color (blue and orange) in correspondence to the regime from Fig.~\ref{fig:phase_diagram}.}
    \label{fig:Artificial_decay}

\end{figure}

Depending on the regularization decay strength $\regdecay$, capacity $\alpha$, source $r$ and noise variance $\sigma^2$, four regimes  can be observed. The derivation of these decays from the asymptotic solution of the Gaussian design problem is sketched in Section \ref{sec:derivation} and detailed in Appendix \ref{appendix:computations}, and here we concentrate on the key results. The different observable decays for the excess error $\epsilon_g-\sigma^2$ are summarized in Fig.~\ref{fig:phase_diagram}, and are given by:  

\begin{itemize}[wide = 0pt,noitemsep]
    \item If $\regdecay\ge \alpha$ (weak regularization $\reg=n^{-\regdecay}$),
    \begin{equation}
        \epsilon_g-\sigma^2=\mathcal{O}\left(\mathrm{max}\left(\sigma^2,\samples ^{-2\alpha\mathrm{min}(r,1)}\right)\right)\, .
    \label{eq:green_red_decay}
    \end{equation}
    The excess error transitions from a fast decay $2\alpha \mathrm{min}(r,1)$ (green region in Fig.~\ref{fig:phase_diagram} and green dashed line in Fig.~\ref{fig:Artificial_noreg}) to a plateau (red region in Fig.~\ref{fig:phase_diagram} and red dashed line in Fig.~\ref{fig:Artificial_noreg}) with no decay as $n$ increases. This corresponds to a crossover from the green region to the red region in the phase diagram Fig.~\ref{fig:phase_diagram}.
    
    \item If $\regdecay \le \alpha$ (strong regularization $\reg=n^{-\regdecay}$) ,
    
    \begin{equation}
        \epsilon_g-\sigma^2=\mathcal{O}\left(\mathrm{max}\left(\sigma^2,n^{1-2\regdecay\rm{min}(r,1)-\frac{\regdecay}{\alpha}}\right)
    \samples ^\frac{\regdecay-\alpha}{\alpha}
    \right).
    \label{eq:blueorange_decay}
    \end{equation}
    The excess error transitions from a fast decay $2\regdecay \mathrm{min}(r,1)$ (blue region in Fig.~\ref{fig:phase_diagram}) to a slower decay $(\alpha-\regdecay)/\alpha$ (orange region in Fig.~\ref{fig:phase_diagram}) as $n$ is increased and the effect of the additive noise kicks in, see Fig.~\ref{fig:Artificial_decay}. The crossover disappears for too slow decays $l\le \alpha/(1+2\alpha\mathrm{min}(r,1))$, as the regularization $\reg$ is always sufficiently large to completely mitigate the effect of the noise. This corresponds to the max in \eqref{eq:blueorange_decay} being realized by its second argument for all $\samples$. 
\end{itemize}

Given these four different regimes as depicted in Fig.~\ref{fig:phase_diagram}, one may wonder about the optimal learning solution when the regularization is fine tuned to its best value. To answer this question, we further define the \textit{asymptotically optimal} regularization decay $\regdecay^\star$ as the value leading to fastest decay of the typical excess error $\epsilon_g-\sigma^2$. We find that two different optimal rates exist, depending on the quantity of data available. 

\begin{figure}
    \centering
    \includegraphics[scale=0.5]{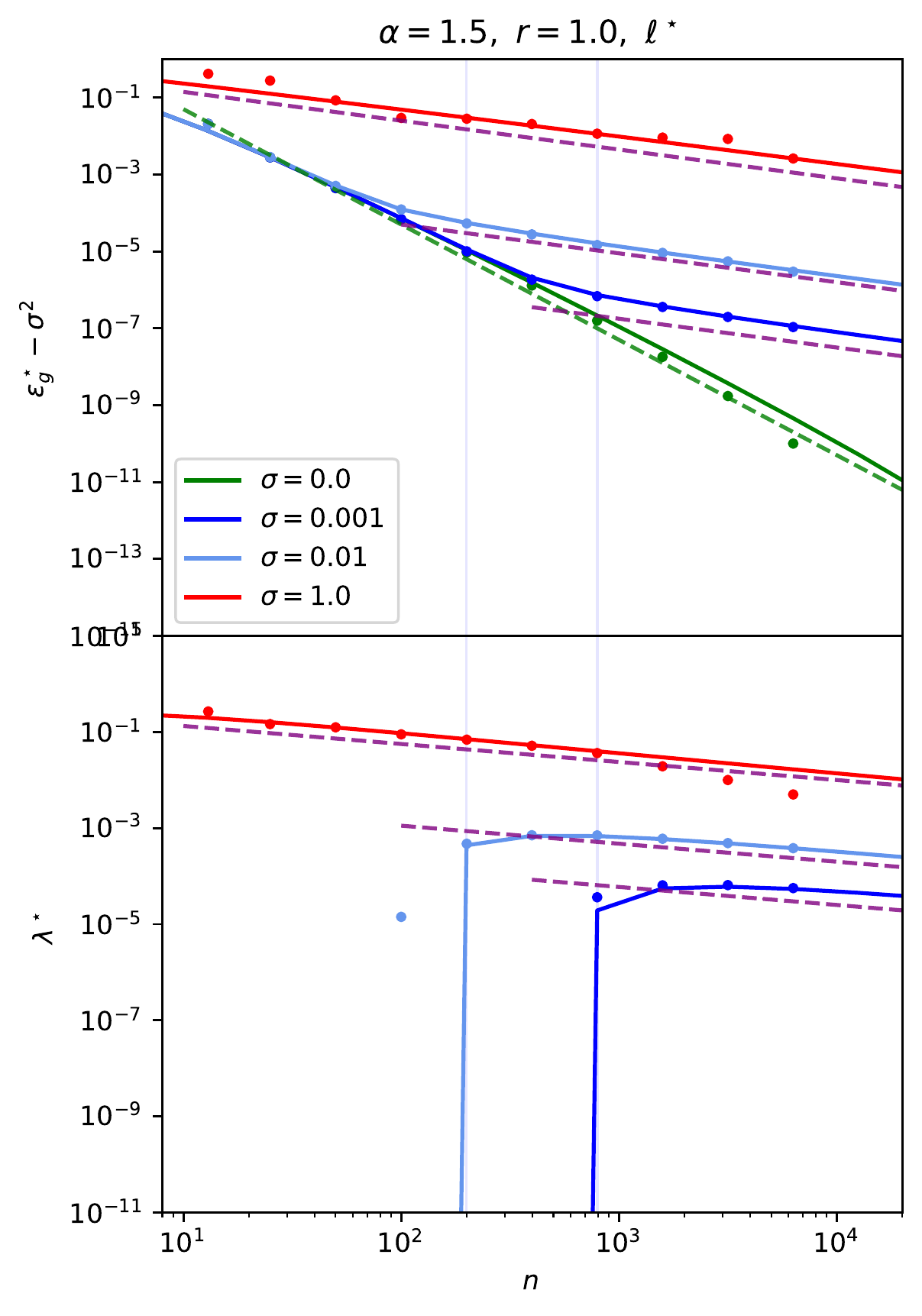}
    \includegraphics[scale=0.5]{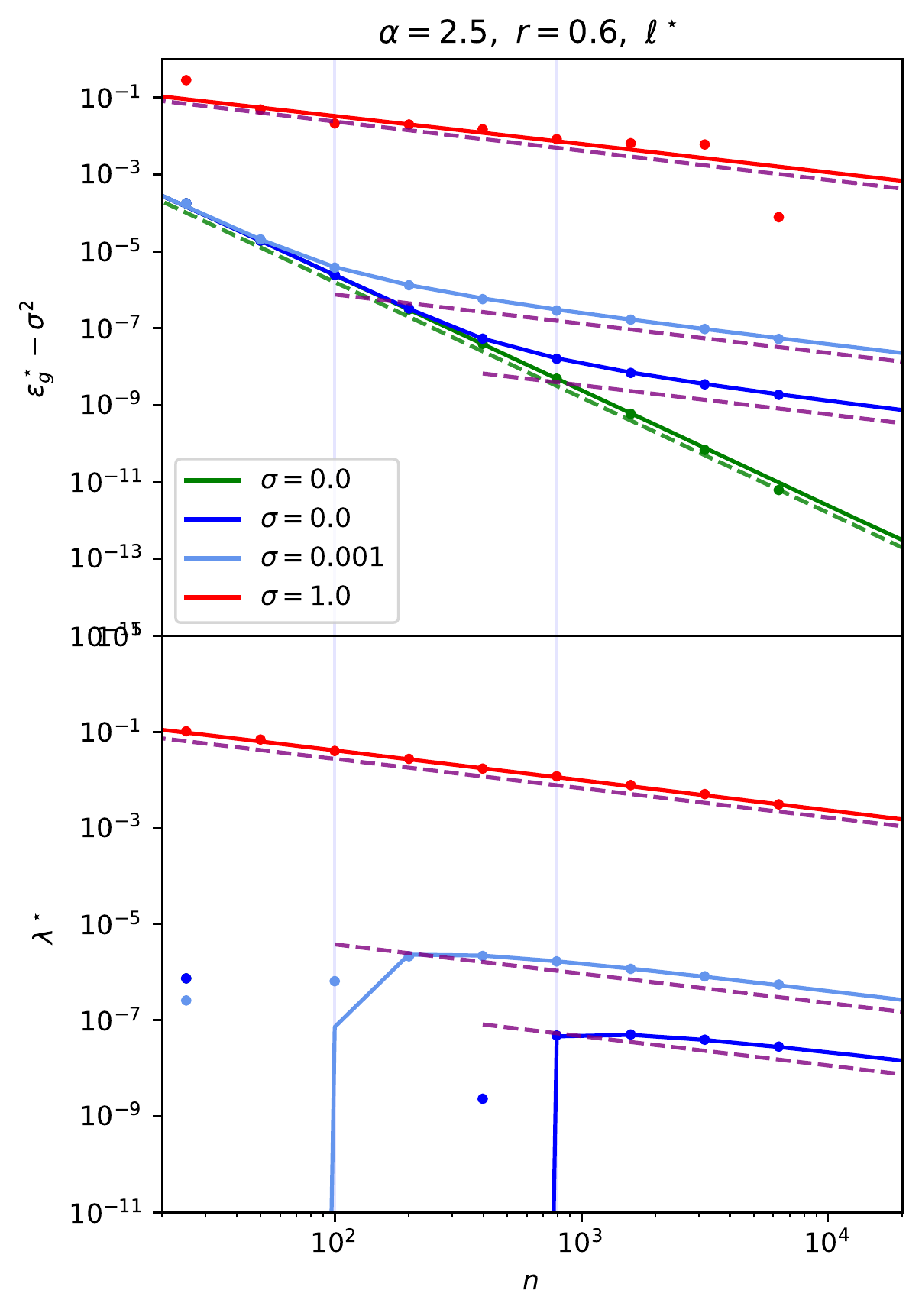}
    \caption{Kernel ridge regression on synthetic data sets with capacity $\alpha$ and source coefficient~$r$. The regularization $\reg$ is chosen as the one minimizing the theoretical prediction for the excess generalization error, deduced from eq.~(\ref{eq:SP_kernel}) using the \texttt{GCM} package associated with \cite{Loureiro2021CapturingTL}. Solid lines correspond to the theoretical prediction of eq.~(\ref{eq:SP_kernel}). Points are simulations conducted with the python \texttt{scikit-learn KernelRidge} package \cite{scikit-learn}, where the feature space dimension has been cut off to $\featuredim=10^4$ for the simulations, and to $10^5$ for the theoretical curves. In simulations, the best $\reg^\star$ was determined using python \texttt{scikit-learn} \texttt{GridSearchCV} cross validation package \cite{scikit-learn}. Note that because cross validation is not adapted to small training sets, a few discrepancies are observed for smaller $\samples$. Dashed lines represent the slopes predicted by theory, with the colors in correspondence to the regimes in Fig.~\ref{fig:phase_diagram}, purple for the purple point in Fig.~\ref{fig:phase_diagram}. Top: excess error. Bottom: optimal $\reg^\star$. Note the noiseless case has $\reg^*=0$.}
    \label{fig:Artificial_CV}

\end{figure}

 \begin{itemize}[wide = 0pt,noitemsep]
     \item If $n \ll n_1^* \approx \sigma^{-\frac{1}{\alpha\mathrm{min}(r,1)}}$, any $\regdecay^\star\in(\alpha,\infty)$ yields excess error decay
     \begin{equation}
         \epsilon_g^\star-\sigma^2\sim \samples^{-2\alpha\mathrm{min}(r,1)}\, .
    \label{eq:opt_noiseless}
     \end{equation}
         \item If $n \gg n_2^* \approx \sigma^{-\mathrm{max}\left(2,\frac{1}{\alpha\mathrm{min}(r,1)}\right)}$,
     \begin{align}
         \epsilon_g^\star-\sigma^2\sim n^{\frac{1}{1+2\alpha\mathrm{min}(r,1)}-1} \, ,&& \text{ by choosing }\quad \reg^\star\sim n^{-\frac{\alpha}{1+2\alpha\mathrm{min}(r,1)}}\, .
    \label{eq:opt_noisy}
     \end{align}
 \end{itemize}
The optimal decay for the excess error $\epsilon_g^\star-\sigma^2$ thus transitions from a fast decay $2\alpha \mathrm{min}(r,1)$ when $n \ll n^*_1$ -- corresponding to, effectively, the optimal rates expected in a "noiseless" situation --  to a slower decay $2\alpha \mathrm{min}(r,1)/(1+2\alpha \mathrm{min}(r,1))$ when $n \gg n^*_2$ corresponding to the classical "noisy" optimal rate, depicted with the purple point in Fig.~\ref{fig:phase_diagram}. This is illustrated in Fig.~\ref{fig:Artificial_CV} where the two rates are observed in succession for the same data as the number of points is increased.

We can now finally clarify the apparent discrepancy in the recent literature discussed in the introduction. The exponent recently reported in \cite{spigler2019asymptotic,bordelon2020} actually corresponds to the "noiseless" regime. In contrast, the rate described in \eqref{eq:opt_noisy} is the classical result \cite{Caponnetto2005FastRF} for the non-saturated case $r<1$ for generic data. We see here that the same rate is also achieved with Gaussian design, and that there are no differences between fixed and Gaussian design as long as the capacity and source condition are matching. We unveiled, however, the existence of two possible sets of optimal rate exponents depending on the number of data samples.

\noindent All setups (effectively non-regularized KRR \eqref{eq:green_red_decay}, effectively regularized KRR \eqref{eq:blueorange_decay} or optimally regularized KRR \eqref{eq:opt_noiseless}, \eqref{eq:opt_noisy}) can therefore exhibit a {\it crossover} from an effectively \textit{noiseless} regime (green or blue in Fig.~\ref{fig:phase_diagram}), to an effectively \textit{noisy} regime (red, orange in Fig.~\ref{fig:phase_diagram}) depending on the quantity of data available. We stress that while the noise is indeed present in the green and blue "noiseless" regimes, its presence is effectively not felt, and noiseless rates are observed. In fact, if the noise is small, one will not observed the classical noisy rates unless an astronomical amount of data is available. This can be intuitively understood as follows:  for small sample size $\samples$, low-variance dimensions are used to overfit the noise, while the spiked subspace of large-variance dimensions is well fitted. In noiseless regions, the excess error is thus characterized by a fast decay.  This phenomenon, where the noise variance is diluted over the dimensions of lesser importance, is connected to the \textit{benign overfitting} discussed by \cite{Bartlett2020BenignOI} and \cite{Tsigler2020BenignOI}. Benign overfitting is possible due to the decaying structure of the co-variance spectrum \eqref{eq:model_def}. As more samples are accessed, further decrease of the excess error requires good generalization also over the low-variance subspace, and the overfitting of the noise results in a slower decay.


While our analysis is for the optimal full-batch learning, we note that a similar crossover in the case of SGD in the effectively non-regularized case  (from green to red) has been discussed in \cite{Berthier2020TightNC,Varre2021LastIC}. It would be interesting to further explore how SGD can behave in the different regimes discussed here. 


When $\reg=\reg_0n^{-\regdecay}$ for a prefactor $\reg_0$ that is allowed to be very small, a \textit{regularization-induced } crossover, similar to the one reported in \cite{bordelon2020}, can also be observed on top of the noise-induced crossover which is the focus of the present work. This setting is detailed in Appendix.~\ref{appendix:Crossover}.

\section{Sketch of the derivation}

\label{sec:derivation}
We provide in this section the main ideas underlying the derivation of the main results exposed in section \ref{section:phase_diagram} and summarized in Fig.~\ref{fig:phase_diagram}. A more detailed discussion is presented in Appendix~\ref{appendix:computations}.

\paragraph{Closed-form solution for Gaussian design ---} Closed-form, rigorous solution of the risk of ridge regression with Gaussian data of arbitrary co-variance in the high-dimensional asymptotic regime have been studied in \cite{dobriban2018high}  \cite{wu2020optimal,richards2021asymptotics}. We shall use here the equivalent notations of \cite{Loureiro2021CapturingTL}, who have the advantage of having rigorous non-asymptotic rates guarantees. Using these characterizations as a starting point, we shall sketch how the crossover phenomena \eqref{eq:green_red_decay} \eqref{eq:blueorange_decay}\eqref{eq:opt_noiseless} and \eqref{eq:opt_noisy}, which are the main contribution of this paper, can be derived. Within the framework of \cite{Loureiro2021CapturingTL}, with high-probability when $n,p$ are large the excess prediction error is expressed as
\begin{equation}
    \epsilon_g-\sigma^2=\rho-2m^{\star}+q^{\star},
\label{eq:Loureiro_generror}
\end{equation}
\noindent with $\rho={\teacher}^{\top}\Sigma\teacher$, and $(m^{\star}, q^{\star})$ are the unique fixed-points of the following self-consistent equations:
\begin{align}
\label{eq:SP_kernel}
    \begin{cases}
        \hat{V} = \frac{\frac{n}{\sdim}}{1+V}\\
        \hat{q} = \frac{n}{\sdim}\frac{\rho+q-2m+\sigma^2}{(1+V)^2}\\
    \end{cases}, && 
    \begin{cases}
		q = \sdim\sum\limits_{\featindex=1}^{p}\frac{\hat{q}\eig_{\featindex}^2+\theta_{\featindex}^{\star 2}\eig_{\featindex}^2 \hat{m}^2}{(\samples\reg+\sdim\hat{V}\eig_{\featindex})^2}\\
		m=\sdim\hat{V}\sum\limits_{\featindex=1}^{p}\frac{\theta_{\featindex}^{\star 2}\eig_{\featindex}^2}{\samples\reg+\sdim\hat{V}\eig_{\featindex}}
	\end{cases}, &&
	\begin{cases}
		V =  \frac{1}{\sdim}\sum\limits_{\featindex=1}^{p}\frac{p\eig_{\featindex}}{\samples\reg + p\hat{V}\eig_{\featindex}}
	\end{cases}
	.
\end{align}
\noindent We recall the reader that $\reg>0$ is the regularisation strength and $\{\eig_{\featindex}\}_{\featindex=1}^{\featuredim}$ are the kernel eigenvalues. The next step is thus to insert the power-law decay \eqref{eq:model_def} for the eigenvalues into \eqref{eq:SP_kernel}, and to take the limit $n,p \to \infty$. We note, however, that this last step is not completely justified rigorously. Indeed,  \cite{dobriban2018high} assumes $p/n=O(1)$ as $n,p\to \infty$ while here we first send $p \to \infty$ and then take the large $n$ limit, thus working effectively with $p/n \to 0$. While the non-asymptotic rates guarantees of \cite{Loureiro2021CapturingTL} are reassuring in this respect, a finer control of the limit would be needed for a fully rigorous justification. Nevertheless, we observed in our experiments that the agreement between theory and numerical simulations for the excess prediction error \eqref{eq:excess} is perfect (see Figs.~\ref{fig:Artificial_noreg}, \ref{fig:Artificial_decay} and \ref{fig:Artificial_CV}). In the large $n$ limit, one can finally close the equation for the excess prediction error into
\begin{equation}
\epsilon_{g}-\sigma^2 = \frac{\sum\limits_{\featindex=1}^{\infty}\frac{\featindex^{-1-2r\alpha}}{\left(1+nz^{-1}\featindex^{-\alpha}\right)^2}}{1-\frac{n}{z^2}\sum\limits_{\featindex=1}^{\infty}\frac{\featindex^{-2\alpha}}{(1+nz^{-1}\featindex^{-\alpha})^2}}+\sigma^2\frac{\frac{ n}{z^2}\sum\limits_{\featindex=1}^{\infty}\frac{\featindex^{-2\alpha}}{(1+nz^{-1}\featindex^{-\alpha})^2}}{1-\frac{n}{z^2}\sum\limits_{\featindex=1}^{\infty}\frac{\featindex^{-2\alpha}}{(1+nz^{-1}\featindex^{-\alpha})^2}}.
\label{eq:decompo}
\end{equation}
with $z$ being a solution of 
\begin{equation}
\label{eq:z}
z\approx \samples\reg+\left(\frac{z}{\samples }\right)^{1-\frac{1}{\alpha}}\int_{\left(\frac{z}{\samples }\right)^{1/\alpha}}^\infty \frac{\dd x}{1+x^\alpha}.
\end{equation}
The detailed derivation is provided in Appendix~\ref{appendix:computations}. We note that this equation was observed with heuristic arguments from statistical physics (using the non-rigorous cavity method) in \cite{Canatar2021SpectralBA}.

The different regimes of excess generalization error rates discussed in Section \ref{section:phase_diagram} are derived from this self-consistent equation. Note that the excess error \eqref{eq:decompo} decomposes over a sum of two contributions, respectively accounting for the sample variance and the noise-induced variance. In contrast to a typical bias-variance decomposition, the effect of the bias introduced in the task for non-vanishing $\reg$ is subsumed in both terms.\\

\paragraph{Derivation of the four regimes ---} If the second term in \eqref{eq:z} dominates, then $z\sim\samples^{1-\alpha}$, which is self consistent if $\regdecay\ge\alpha$. This is the \textit{effectively non-regularized regime}, where the regularization $\reg$ is not sensed, and corresponds to the green and red regimes in the phase diagram in Fig.~\ref{fig:phase_diagram}. This scaling of $z$ can then be used to estimate the asymptotic behaviour of the sample and noise induced variance in the decomposition on the excess error \eqref{eq:decompo},
yielding
\begin{equation}
    \epsilon_g-\sigma^2=\mathcal{O}(n^{-2\alpha\mathrm{min}(r,1)})+\sigma^2\mathcal{O}(1),
\label{eq:non-regularized}
\end{equation}
which can be rewritten more compactly as \eqref{eq:green_red_decay}. Therefore, for small sample sizes the sample variance drives the decay of the excess prediction error, while for larger samples sizes the noise variance dominates and causes the error to plateau. The crossover happens when both variance terms in \eqref{eq:non-regularized} are balanced, around
\begin{equation}
    n\sim\sigma^{-\frac{1}{\alpha\mathrm{min}(r,1)}},
    \label{eq:verticalline}
\end{equation}
which corresponds to the vertical part of the crossover line in Fig.~\ref{fig:phase_diagram}.

If the first term $n\reg$ dominates in \eqref{eq:z}, then $z\sim n\reg$, which is consistent provided that $\regdecay<\alpha$. This is the \textit{effectively regularized regime} (blue, orange regions in Fig.~\ref{fig:phase_diagram}). The two variances in \eqref{eq:decompo} are found to asymptotically behave like
\begin{equation}
    \epsilon_g-\sigma^2=\mathcal{O}(n^{-2\regdecay\mathrm{min}(r,1)})+\sigma^2\mathcal{O}(n^{\frac{\regdecay-\alpha}{\alpha}}),
\label{eq:Regularized}
\end{equation}
which can be rewritten more compactly as \eqref{eq:blueorange_decay}. If the decay of the noise variance term $(\alpha-\regdecay)/\alpha$ is faster than the $2\regdecay\mathrm{min}(r,1)$ decay of the sample variance term, then the latter always dominates and no crossover is observed. This is the case for $\regdecay<\alpha/(1+2\alpha\mathrm{min}(r,1))$. If on the contrary the decay of the noise variance term is the slowest, then this term dominates at larger $\samples$, with a crossover when both terms in \eqref{eq:Regularized} are balanced, around
\begin{equation}
n\sim\sigma^{\frac{2}{1-\frac{\regdecay}{\alpha}(1+2\alpha\mathrm{min}(r,1))}}
\label{eq:curveline}
\end{equation}
Eqs. \eqref{eq:non-regularized} and \eqref{eq:Regularized} are respectively equivalent to \eqref{eq:green_red_decay} and \eqref{eq:blueorange_decay}, and completely define the four regimes observable in Fig.~\ref{fig:phase_diagram}. Equations \eqref{eq:curveline} and \eqref{eq:verticalline} give the expression for the crossover line in Fig.~\ref{fig:phase_diagram}.

\paragraph{Asymptotically optimal regularization ---}
Determining the asymptotically optimal $\regdecay^\star$ is a matter of finding the $\regdecay$ leading to fastest excess error decay. We focus on the far left part and the far right part of the phase diagram Fig.~\ref{fig:phase_diagram}.

In the $ n \gg  n_2^\star\approx\sigma^{-\mathrm{max}\left(2,\frac{1}{\alpha\mathrm{min}(r,1)}\right)}$ limit where the crossover line confounds itself with its $\regdecay=\alpha/(1+2\alpha\mathrm{min}(r,1))$ asymptot, this is tantamount to solving the maximization problem 
\begin{align}
    \regdecay^\star=\underset{\regdecay}{\mathrm{argmax}}\left(
    2\regdecay\mathrm{min}(r,1)\mathbbm{1}_{0<\regdecay<\frac{\alpha}{(1+2\alpha\mathrm{min}(r,1))}}+\frac{\alpha-\regdecay}{\alpha}\mathbbm{1}_{\frac{\alpha}{(1+2\alpha\mathrm{min}(r,1))}<\regdecay<\alpha}+0\times \mathbbm{1}_{\alpha<\regdecay}
    \right)
\end{align}
which admits as solution \eqref{eq:opt_noisy}. In the $\samples\ll n_1^\star\approx\sigma^{-\frac{1}{\alpha\mathrm{min}(r,1)}}$ range, the maximization of the excess error decay reads
\begin{align}
    \regdecay^\star=\underset{\regdecay}{\mathrm{argmax}}\left(
    2\regdecay\mathrm{min}(r,1)\mathbbm{1}_{0<\regdecay<\alpha}+2\alpha\mathrm{min}(r,1)\mathbbm{1}_{\alpha<\regdecay}
    \right),
\end{align}
and admits as solution \eqref{eq:opt_noiseless}.\\


\section{Illustration on simple real data sets}

\label{section:real_data}

\begin{figure}[ht]
    \centering
    \includegraphics[scale=0.5]{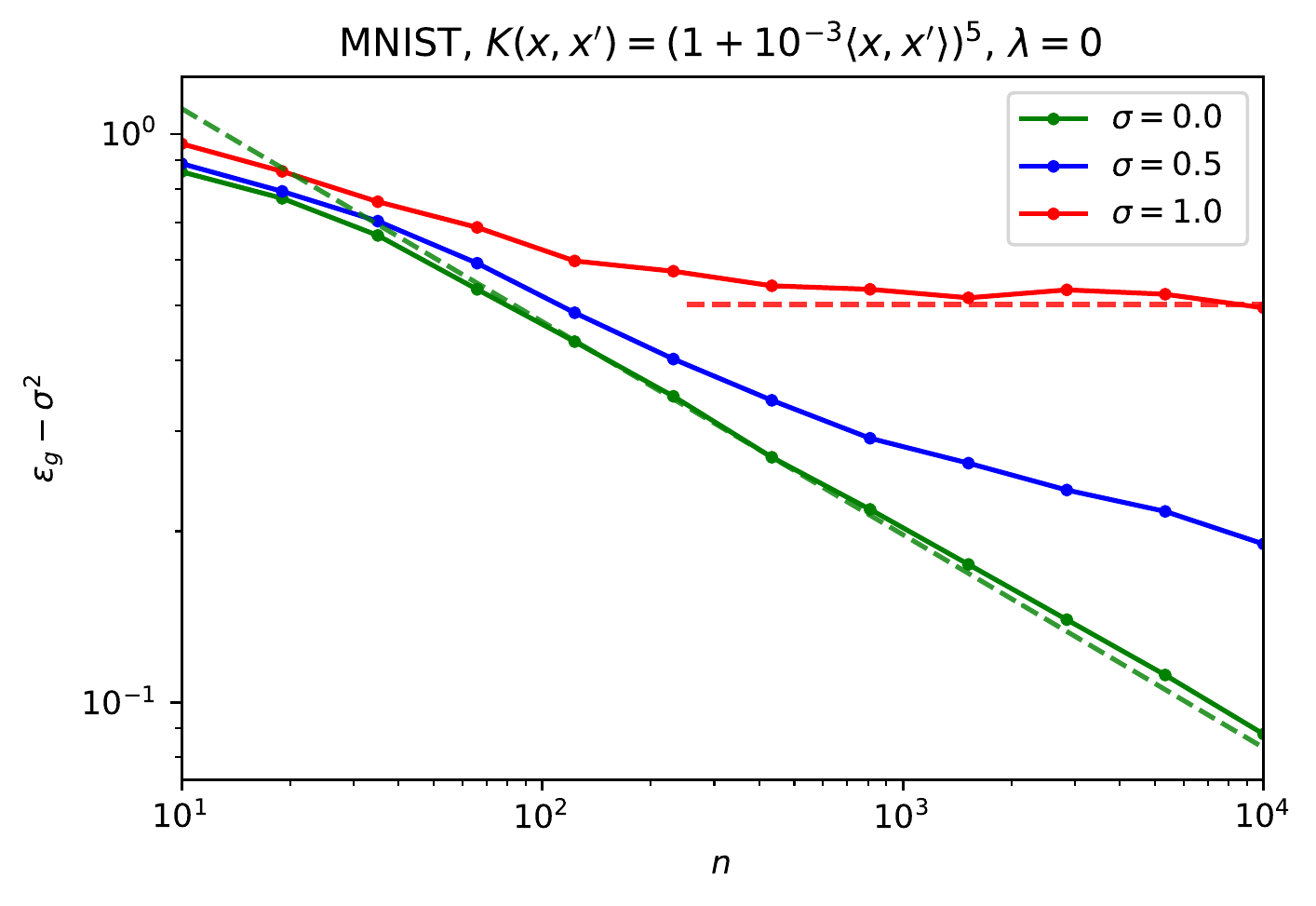}
    \includegraphics[scale=0.5]{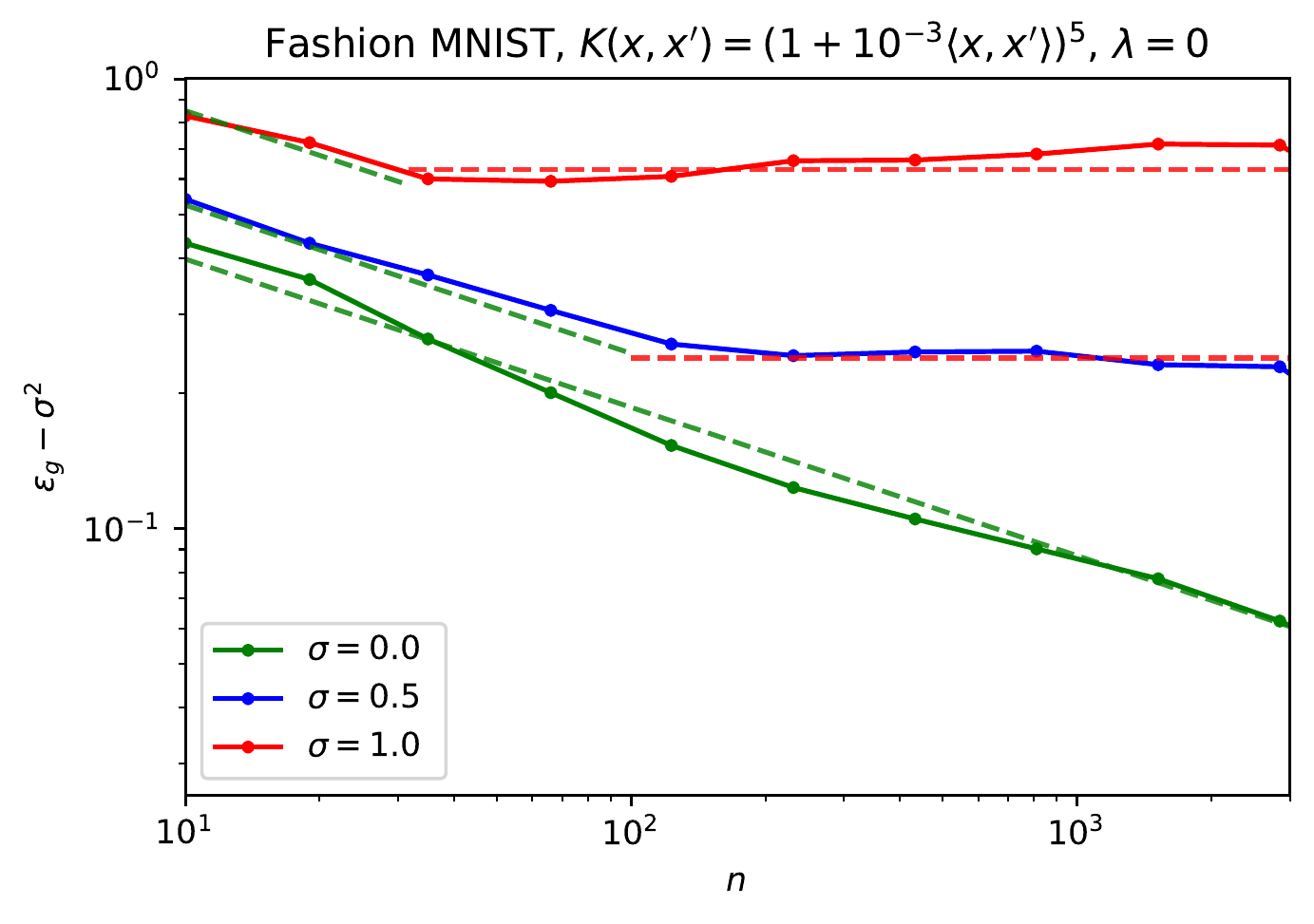}
    \caption{Excess error for MNIST odd versus even (above) and Fashion MNIST t-shirt versus coat (below) with labels corrupted by noise of variance $\sigma^2$. The kernel used is indicated in the title. Solid lines with points come from numerical experiments with zero regularization. Dashed lines are the slopes $-2\alpha r$ (as $r<1$) or $0$, predicted by the theory from the empirical values of $\alpha,r$ measured from the Gram matrix spectrum and the teacher for each data set, see Table \ref{table:measuredalphar}. Colors of the dashed lines (green \& red) indicate the regimes in Fig.~\ref{fig:phase_diagram}.}
    \label{fig:Real_noreg}

\end{figure}

\begin{figure}[t]
    \centering
    \includegraphics[scale=0.5]{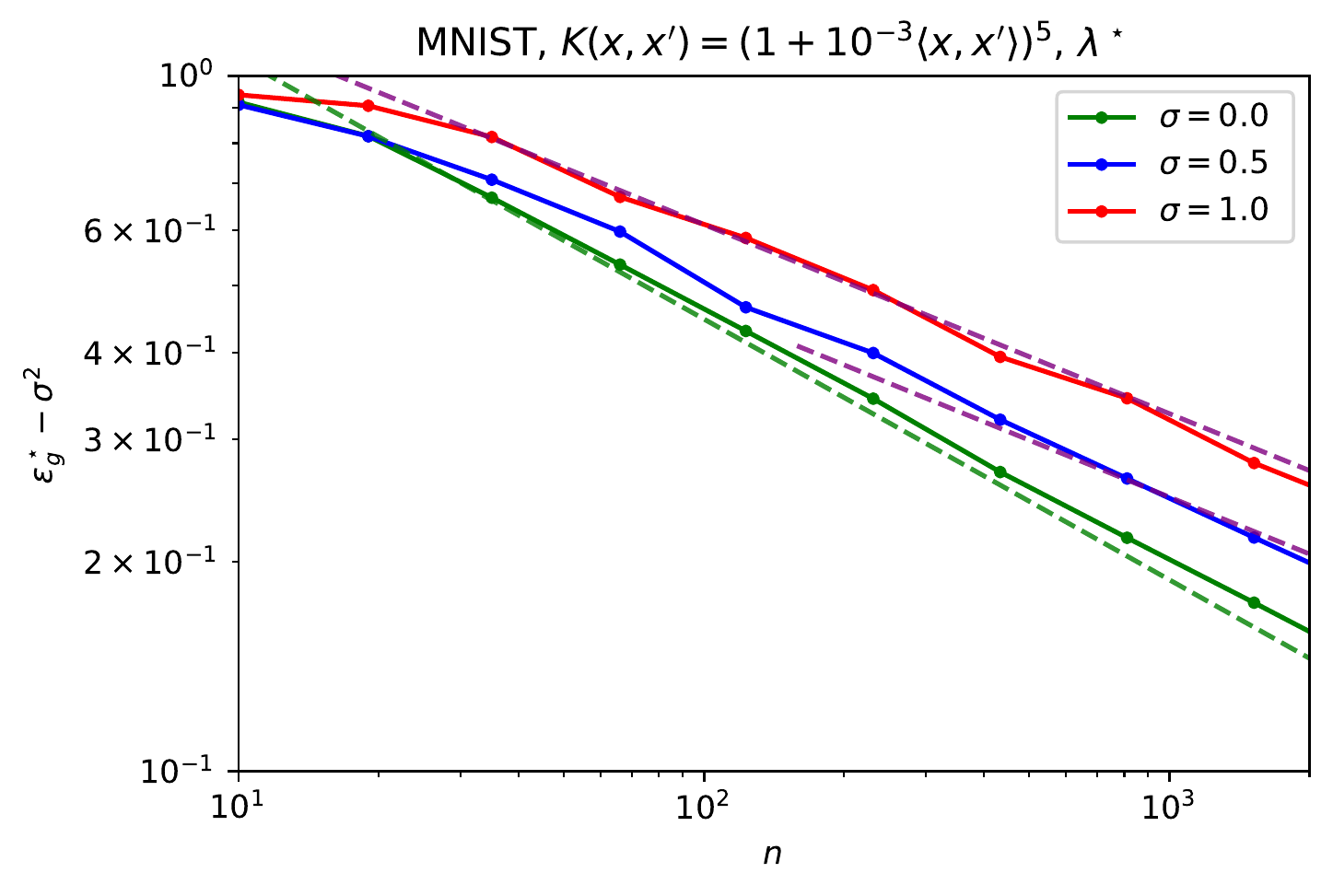}
    \includegraphics[scale=0.5]{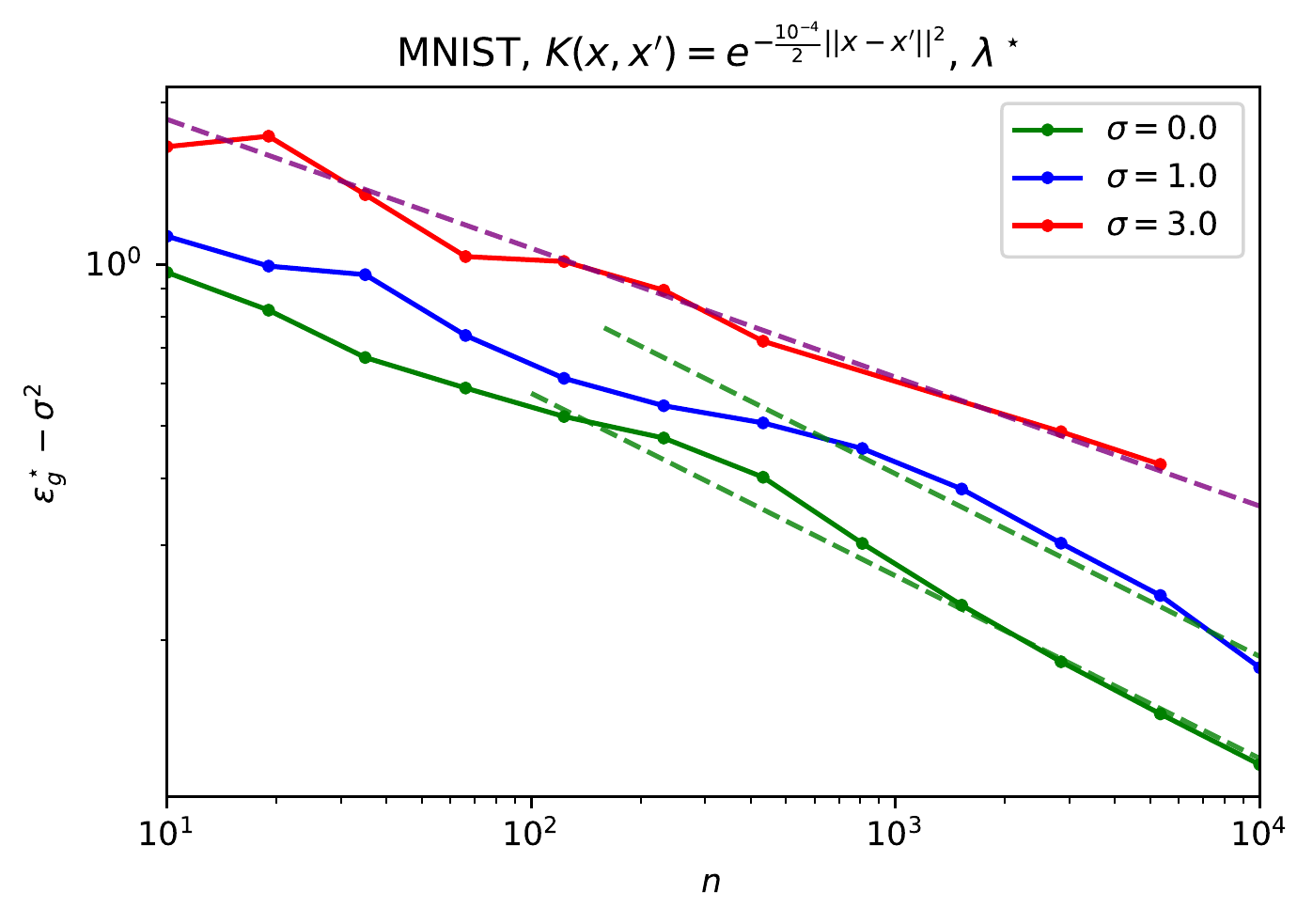}
    \includegraphics[scale=0.5]{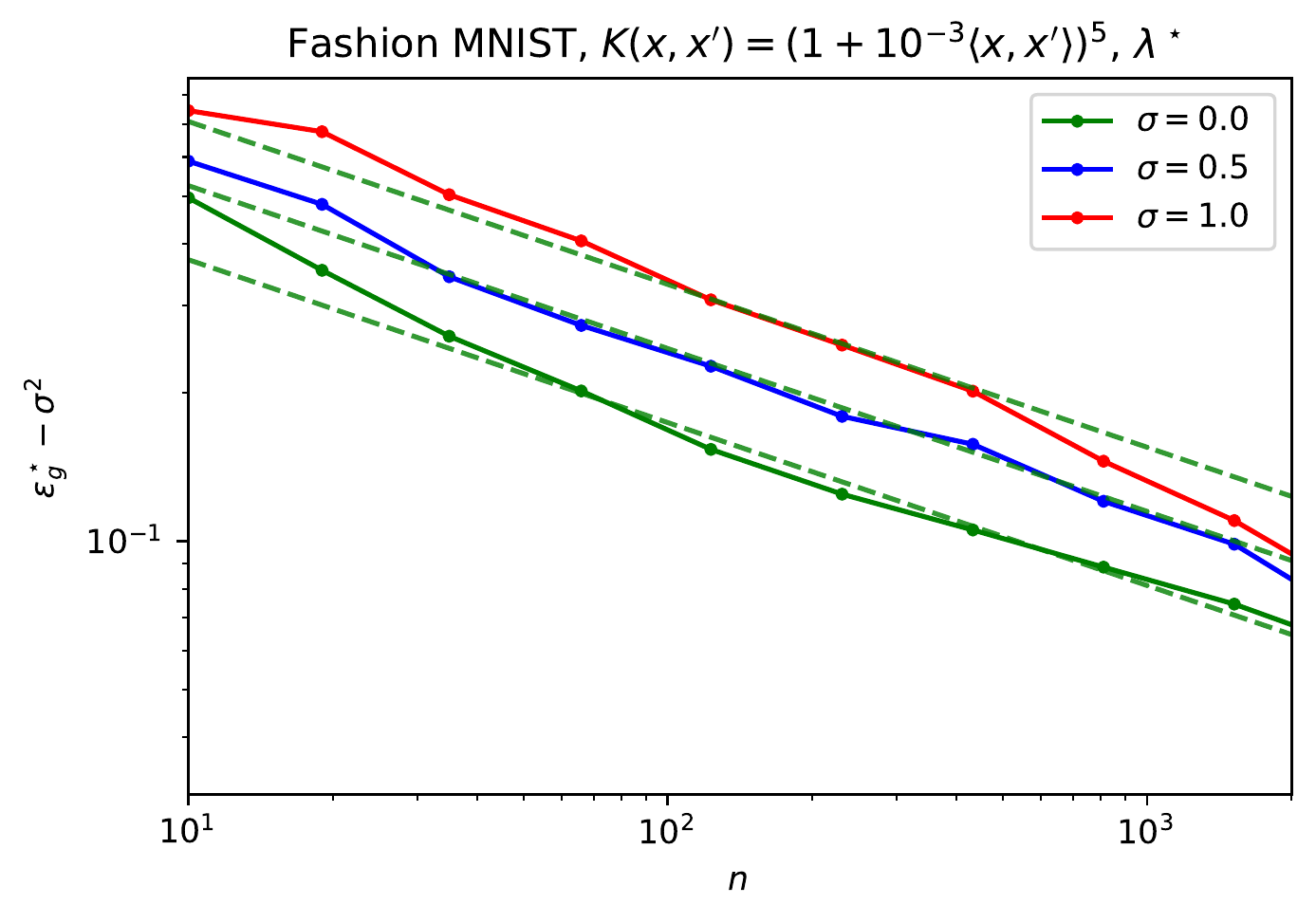}
    \includegraphics[scale=0.5]{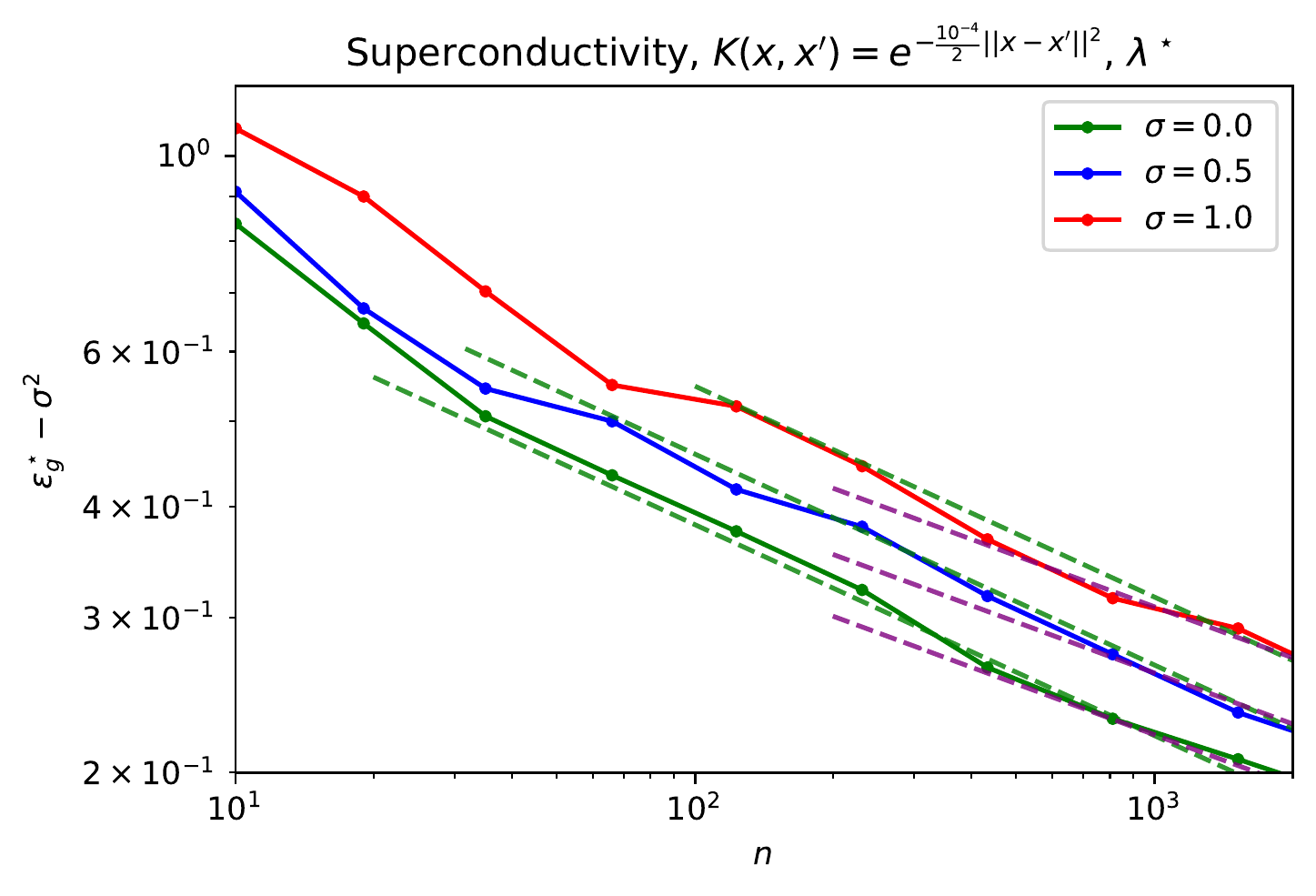}
    \caption{Excess error for MNIST odd versus even, and Fashion MNIST t-shirt versus coat, and the critical temperature regression. The kernel used is indicated in the title. Solid lines with dots come from numerical experiments with the regularization optimized using the python \texttt{scikit-learn} \texttt{GridSearchCV} package \cite{scikit-learn}. Dashed lines are the slopes predicted by the theory, from the empirical values of $\alpha,r$ measured from the Gram matrix spectrum and the teacher for each data set, see Table \ref{table:measuredalphar}. Colors of the dashed lines indicate the regime in Fig.~\ref{fig:phase_diagram}. 
    }
    \label{fig:Real_CV}
\end{figure}

In this section we show that the derived decay rates can indeed be observed in real data sets with labels artificially corrupted by additive Gaussian noise. For real data, the decay model in eq.~\eqref{eq:model_def} is idealized, and in practice there is no firm reason to expect a power-law decay. However, we do find that for some of the data sets and kernels we investigated, the power law fit is reasonable and can be used to estimate the exponents $\alpha$ and $r$, see Appendix \ref{appendix:real_data} for details. For those cases, we compare the theoretically predicted exponents, eqs.~\eqref{eq:green_red_decay}, \eqref{eq:blueorange_decay}, \eqref{eq:opt_noiseless} and \eqref{eq:opt_noisy} with the empirically measured learning curve, and obtain a very good agreement. We stress that the decay rates are not obtained by fitting the learning curves, but rather by fitting the exponents $\alpha$ and $r$ from the data. We also observe the crossover from the noiseless (blue, green in Fig.~\ref{fig:phase_diagram}) to the noisy (orange, red in Fig.~\ref{fig:phase_diagram}) regime given by the theory. 
Here we illustrate this with the learning curves for the following three data sets:
\begin{itemize}[wide = 0pt,noitemsep]
    \item MNIST even versus odd, a data set of $7\times 10^4$ $28\times 28$ images of handwritten digits. Even (odd) digits were assigned label $y=1+\sigma\mathcal{N}(0,1)$ ($y=-1+\sigma\mathcal{N}(0,1)$).
    \item Fashion MNIST t-shirts versus coats, a data set of $14702$ $28\times 28$ images of clothes from an online shopping platform \cite{xiao2017}. T-shirts (coats) were assigned label $y=1+\sigma\mathcal{N}(0,1)$ ($y=-1+\sigma\mathcal{N}(0,1)$).
    \item Superconductivity \cite{Hamidieh2018ADS}, a data set of 81 attributes of 21263 superconducting materials. The target $y^\mu$ corresponds to the critical temperature of the material, corrupted by additive Gaussian noise.
\end{itemize}
Learning curves are illustrated for a radial basis function (RBF) kernel $K(x,x')=e^{-\frac{\gamma}{2}||x-x'||^2}$ with parameter $\gamma=10^{-4}$ and a degree $5$ polynomial kernel $K(x,x')=(1+\gamma \langle x,x'\rangle)^5$ with parameter $\gamma=10^{-3}$. In Fig.~\ref{fig:Real_noreg} the regularization $\reg$ was set to $0$, while in Fig.~\ref{fig:Real_CV} $\reg$ was optimized for each sample size $\samples$ using the python \texttt{scikit-learn} \texttt{GridSearchCV} package \cite{scikit-learn}. KRR was carried out using the \texttt{scikit-learn KernelRidge} package \cite{scikit-learn}. The values of $\alpha, r$ were independently measured (see Appendix \ref{appendix:real_data}) for each data set, and the estimated values summarized in Table~\ref{table:measuredalphar}. From these values the theoretical decays \eqref{eq:green_red_decay},  \eqref{eq:opt_noiseless} and \eqref{eq:opt_noisy} were computed, and compared with the simulations with very good agreement. Since for real data the power-law form \eqref{eq:model_def} does not exactly hold (see Fig.~\ref{fig:Measure_params} in the appendix), the estimates for $\alpha, r$ slightly vary depending on how the power-law is fitted. The precise procedure employed is described in Appendix \ref{appendix:real_data}. Overall this variability does not hurt the good agreement with the simulated learning curves in Fig.~\ref{fig:Real_noreg} and \ref{fig:Real_CV}.

When $\reg=0$  (Fig.~\ref{fig:Real_noreg}) the characteristic plateau for large label noises is observed for both MNIST \& Fashion MNIST. For polynomial kernel regression on Fashion MNIST (Fig.~\ref{fig:Real_noreg} right), the crossover between noiseless (slope $-2\alpha r$ as $r<1$) and noisy (slope $0$) regimes is apparent on the same learning curve at noise levels $\sigma=0.5, 1$. For MNIST, the $\sigma=0$ ($\sigma=1$) curve is in the noiseless (noisy) regime for larger $n$, while at intermediary noise $\sigma=0.5$, and small $n$ for $\sigma=1$, the curve is in the crossover regime between noiseless and noisy, consequently displaying in-between decay. Our results for the decays for $\sigma=0$ agree with simulations for RBF regression on MNIST provided in \cite{spigler2019asymptotic}.

For optimal regularization $\reg=\reg^\star$ (Fig.~\ref{fig:Real_CV}), as the measured $r<1$ we have exponents $-2r\alpha$ for the noiseless regime and $-2r\alpha/(1+2r\alpha)$ for noisy. Since the measured value of $2r\alpha$ is rather small the difference between the two rates is less prominent. Nevertheless, it seems that in our experiments the noisy regime is observed for polynomial and RBF kernels on MNIST and 
$\sigma=0.5, 1$. For Superconductivity, the green and purple decay have close values and it is difficult to clearly identify the regime. For Fashion MNIST only the noiseless rate is observable in the considered noise range and sample range.

\begin{table}[]
    \centering
    \begin{tabular}{ |p{3cm}|p{6cm}||p{1cm}|p{1cm}|  }
 \hline
 Dataset& Kernel & $\alpha$&$r$\\
 \hline
 Fashion MNIST & $K(x,x')=(1+10^{-3}\langle x,x'\rangle)^5$&$1.3$    &$0.13$\\
 MNIST & $K(x,x')=(1+10^{-3}\langle x,x'\rangle)^5$&$1.2$    &$0.15$\\
 MNIST & $K(x,x')=\mathrm{exp}(-10^{-4}||x-x'||^2/2)$&$1.65$    &$0.097$\\
  Superconductivity & $K(x,x')=\mathrm{exp}(-10^{-4}||x-x'||^2/2)$&$2.7$    &$0.046$\\
\hline
\end{tabular}
\caption{Values of the source and capacity coefficients \eqref{eq:source_capa_conditions} as estimated from the data sets. The details on the estimation procedure can be found in Appendix \ref{appendix:real_data}. }
    \label{table:measuredalphar}

\end{table}

\section*{Conclusion}
To conclude, we unify hitherto disparate lines of work, and give a comprehensive study of observable regimes, along the associated decay rates for the excess error, for kernel ridge regression with features having power-law co-variance spectrum. We show that the effect of the noise only kicks in at larger sample complexity, meaning, in particular, that the KRR transitions from a \textit{noiseless} regime with fast error decay to a \textit{noisy} regime with slower decay. This crossover is shown to happen for zero, decaying and optimized regularization, and is observed on a variety of real data sets corrupted with label noise. 


\section*{Acknowledgements}
We warmly thank Loucas Pillaud-Vivien for discussions, and for his help in navigating the literature on kernel ridge regression. We also thank Matthieu Wyart and Cedric Gerbelot for discussions. We acknowledge funding from the ERC under the European Union's Horizon 2020 Research and Innovation Programme Grant Agreement 714608-SMiLe, and from the French National Research Agency grants ANR-17-CE23-0023-01 PAIL.

\medskip
\bibliographystyle{unsrt}
\bibliography{main.bib}


\newpage
\appendix

\section{Derivation of the decays}
\label{appendix:computations}
\subsection{Equations for Gaussian design}
In this Appendix we discuss the derivation of eqs.~\eqref{eq:Loureiro_generror}, \eqref{eq:SP_kernel} describing the excess prediction error for the ridge regression problem with generic covariance matrix. Exact asymptotic formulas for the excess prediction error of least-squares and ridge regression are a classic result in high-dimensional statistics, and have been derived in many different works \cite{karoui2013asymptotic, dobriban2018high, thrampoulidis2018precise, hastie2019surprises}. In this manuscript, we follow the presentation given in \cite{Loureiro2021CapturingTL}, which is particularly adapted to our derivation and has the advantage to hold rigorously at large but finite number of samples $n$ and features $p$. 

\label{subsection:app:mapping}
We start by reviewing the formulas in \cite{Loureiro2021CapturingTL}. Consider the ridge regression problem on $\samples$ independent $\sdim$-dimensional samples $\{u^\mu,y^\mu\}_{\mu=1}^\samples$, defined by a minimisation of the following empirical risk:
\begin{equation}
    \hat{\mathcal{R}}_{n}(w)= \sum\limits_{\mu=1}^n \left(\frac{w\cdot u^\mu}{\sqrt{\sdim}}-y^\mu\right)^2+ \lambda ||w||_2^2.
\label{eq:other_risk}
\end{equation}
Assume a Gaussian design $u^\mu\overset{d}{=}\mathcal{N}(0,\Sigma)$ with diagonal covariance $\Sigma=\mathrm{diag}(\eig_1,...,\eig_p)$ and labels $y^\mu$ generated from a teacher/target/oracle $\theta^\star\in\mathbb{R}^\sdim$:
\begin{equation}
    y^\mu=\frac{\theta^\star\cdot u^\mu}{\sqrt{\sdim}}+\sigma\mathcal{N}(0,1).
\end{equation}
Under the assumptions
\begin{enumerate}[font={\bfseries},label={(A\arabic*)}]
\item $n\gg 1,\sdim\gg 1, \samples/\sdim=\mathcal{O}(1)$,
\item $0<||\theta^\star||^2/\sdim<\infty$,
\end{enumerate}
there exists constants $C,c,c'>0$ such that for all $0<\epsilon<c'$,
\begin{equation}
    \mathbb{P}\left(|\epsilon_g-\sigma^2-(\rho-2m^{\star}+q^{\star})|>\epsilon\right)<\frac{C}{\epsilon}e^{-cn\epsilon^2}.
\label{eq:app:cedric}
\end{equation}
\noindent where $\rho=\theta^\star\cdot\Sigma\cdot \theta^\star/p$, and $(m^{\star},q^{\star})$ are fixed-points of the following \textit{self-consistent equations}
\begin{align}
\label{eq:app:SP_kernel_notrescaled}
    \begin{cases}
        \hat{V} = \hat{m} = \frac{\frac{n}{\sdim}}{1+V}\\
        \hat{q} = \frac{n}{\sdim}\frac{\rho+q-2m+\sigma^2}{(1+V)^2}\\
    \end{cases}, && 
    \begin{cases}
		V =  \frac{1}{\sdim}\sum\limits_{\featindex=1}^{p}\frac{\eig_{\featindex}}{\lambda + \hat{V}\eig_{\featindex}}\\
		q = \frac{1}{\sdim}\sum\limits_{\featindex=1}^{p}\frac{\hat{q}\eig_{\featindex}^2+\theta_{\featindex}^{\star 2}\eig_{\featindex}^2 \hat{m}^2}{(\lambda+\hat{V}\eig_{\featindex})^2}\\
		m=\frac{\hat{m}}{\sdim}\sum\limits_{\featindex=1}^{p}\frac{\theta_{\featindex}^{\star 2}\eig_{\featindex}^2}{\lambda+\hat{V}\eig_{\featindex}}
	\end{cases}.
\end{align}
Note that the risk considered in eq.~\eqref{eq:other_risk} slightly differs from eq.~\eqref{eq:KRR_risk} by: a) a $1/n$ factor multiplying the sum, b) additional $\sqrt{p}$ scalings and c) the fact that it is written for finite $p$. %
Accounting for these differences, we can rewrite Theorem 1 of \cite{Loureiro2021CapturingTL} in our setting as:
\begin{equation}
    \epsilon_g-\sigma^2=\underset{p\rightarrow\infty}{\lim}(\rho-2m^{\star}+q^{\star}),
\label{eq:app:Loureiro_generror}
\end{equation}
with $\rho={\teacher}^{\top}\Sigma\teacher$, and $(m^{\star},q^{\star})$ fixed-points of
\begin{align}
\label{eq:app:SP_kernel}
    \begin{cases}
        \hat{V} = \frac{\frac{n}{\sdim}}{1+V}\\
        \hat{q} = \frac{n}{\sdim}\frac{\rho+q-2m+\sigma^2}{(1+V)^2}\\
    \end{cases}, && 
    \begin{cases}
		V =  \frac{1}{\sdim}\sum\limits_{\featindex=1}^{p}\frac{p\eig_{\featindex}}{\samples\reg + p\hat{V}\eig_{\featindex}}\\
		q = \sdim\sum\limits_{\featindex=1}^{p}\frac{\hat{q}\eig_{\featindex}^2+\theta_{\featindex}^{\star 2}\eig_{\featindex}^2 \hat{m}^2}{(\samples\reg+\sdim\hat{V}\eig_{\featindex})^2}\\
		m=\sdim\hat{V}\sum\limits_{\featindex=1}^{p}\frac{\theta_{\featindex}^{\star 2}\eig_{\featindex}^2}{\samples\reg+\sdim\hat{V}\eig_{\featindex}}
	\end{cases}.
\end{align}
Note, however, that rescaling from \eqref{eq:app:SP_kernel_notrescaled} to \eqref{eq:app:SP_kernel}, sending $p\rightarrow\infty$ while keeping $\samples$ finitely large, and further allowing $\reg$ to scale with $\samples$ all break the initial assumptions of Theorem 1 \cite{Loureiro2021CapturingTL}, thereby losing the control in eq.~\eqref{eq:app:cedric}. Therefore, strictly speaking the results derived hereafter are not rigorous, and we assume that the typical excess error can still be computed from eq.~\eqref{eq:app:Loureiro_generror}. In fact, this is well-justified by comparing the results obtained from extrapolating the theory with finite instance simulation, e.g. Figs. \ref{fig:Artificial_noreg}, \ref{fig:Artificial_decay}, and \ref{fig:Artificial_CV}.

\subsection{Self-consistent equations for the excess prediction error}
\label{subsection:app:selfconsistent}
Defining $z = \frac{n^2}{\sdim} \frac{\lambda}{\hat{V}}$, the equations \eqref{eq:app:SP_kernel} allow to write

\begin{equation}
\label{eq:app:z}
z = n\lambda+\frac{z}{n}\sum\limits_{\featindex=1}^{p}\frac{\eig_{\featindex}}{\frac{z}{n }+\eig_{\featindex}}.
\end{equation}

An expression for the excess error $\epsilon_g-\sigma^2$ can be obtained combining \eqref{eq:app:Loureiro_generror} with \eqref{eq:app:SP_kernel}:
\begin{align}
\epsilon_{g}-\sigma^2 &\underset{(a)}{=}\underset{p\rightarrow\infty}{\lim} \frac{1}{\sdim}\sum\limits_{\featindex=1}^{\sdim}	\left[\theta_{\featindex}^{\star 2}\sdim\eig_{\featindex}+\frac{\hat{q}\sdim^2\eig_{\featindex}^2+\theta_{\featindex}^{\star 2}\sdim^2\eig_{\featindex}^2\hat{m}^{2}}{(n\lambda+\hat{V}\sdim\eig_{\featindex})^2} - \frac{2\hat{m}\theta_{\featindex}^{\star 2}\sdim^2\eig_{\featindex}^2}{n\lambda+\hat{V}\sdim\eig_{\featindex}}\right]\notag\\
&\underset{(b)}{=} \underset{p\rightarrow\infty}{\lim}\sum\limits_{\featindex=1}^{\sdim}\frac{\theta_{\featindex}^{\star 2}\eig_{\featindex}\left(n\lambda+\hat{V}\sdim\eig_{\featindex}\right)^2+\frac{\sdim^2}{\samples}\eig_{\featindex}^2\hat{V}^{2}\epsilon_{g}+\hat{V}^{2}\theta_{\featindex}^{\star 2}\sdim\eig_{\featindex}^2-2\theta_{\featindex}^{\star 2}\hat{V}\sdim\eig_{\featindex}^2\left(n\lambda+\hat{V}\sdim\eig_{\featindex}\right)}{(n\lambda+\hat{V}\sdim\eig_{\featindex})^2}\noindent\\
&= \underset{p\rightarrow\infty}{\lim}\sum\limits_{\featindex=1}^{\sdim}\frac{\frac{\sdim^2}{\samples}\eig_{\featindex}^2\hat{V}^{2}\epsilon_{g}+n^2\lambda^2\theta_{\featindex}^{\star 2}\eig_{\featindex}}{(n\lambda+\hat{V}\sdim\eig_{\featindex})^2},
\end{align}
\noindent thus
\begin{equation}
\label{eq:app:kernelridge_gen_error}
\epsilon_{g} = \underset{p\rightarrow\infty}{\lim}\frac{\frac{z^2 }{n ^2 }\sum\limits_{\featindex=1}^{\sdim}\frac{\theta_{\featindex}^{\star 2}\eig_{\featindex}}{\left(z\frac{1}{n}+\eig_{\featindex}\right)^2}+\sigma^2}{1-\frac{1}{n}\sum\limits_{\featindex=1}^{\sdim}\frac{\eig_{\featindex}^2}{(z\frac{1}{n}+\eig_{\featindex})^2}}.
\end{equation}
Therefore, for the excess prediction error:
\begin{equation}
\label{eq:app:kernelridge_gen_error2}
\epsilon_{g}-\sigma^2 = \underset{p\rightarrow\infty}{\lim}\frac{\frac{z^2 }{n ^2 }\sum\limits_{\featindex=1}^{\sdim}\frac{\theta_{\featindex}^{\star 2}\eig_{\featindex}}{\left(z\frac{1}{n}+\eig_{\featindex}\right)^2}+\frac{\sigma^2}{n}\sum\limits_{\featindex=1}^{\sdim}\frac{\eig_{\featindex}^2}{(z\frac{1}{n}+\eig_{\featindex})^2}}{1-\frac{1}{n}\sum\limits_{\featindex=1}^{\sdim}\frac{\eig_{\featindex}^2}{(z\frac{1}{n}+\eig_{\featindex})^2}}.
\end{equation}

We now assume power-law form for the covariance spectrum and the teacher coordinates \eqref{eq:model_def}
\begin{align}
\label{eq:app:decayansatz}
\eig_{\featindex}=\featindex^{-\alpha}, && \theta_{\featindex}^{\star 2}\eig_{\featindex}=\featindex^{-1-2r\alpha},
\end{align}

Then equation \eqref{eq:app:kernelridge_gen_error} can be simplified to
\begin{equation}
\epsilon_{g}-\sigma^2 = \underset{p\rightarrow\infty}{\lim}\frac{\frac{z^2 }{n ^2 }\sum\limits_{\featindex=1}^{\sdim}\frac{\featindex^{-1-2r\alpha}}{\left(z\frac{1}{n}+\featindex^{-\alpha}\right)^2}+\frac{\sigma^2}{n}\sum\limits_{\featindex=1}^{\sdim}\frac{\featindex^{-2\alpha}}{(z\frac{1}{n}+\featindex^{-\alpha})^2}}{1-\frac{1}{n}\sum\limits_{\featindex=1}^{\sdim}\frac{\featindex^{-2\alpha}}{(z\frac{1}{n}+\featindex^{-\alpha})^2}},
\end{equation}
which has a meaningful limit as $\sdim\rightarrow \infty$ (with $n$, $\lambda$ kept fixed):
\begin{equation}
\epsilon_{g}-\sigma^2 = \frac{\sum\limits_{\featindex=1}^{\infty}\frac{\featindex^{-1-2r\alpha}}{\left(1+nz^{-1}\featindex^{-\alpha}\right)^2}+\frac{\sigma^2 n}{z^2}\sum\limits_{\featindex=1}^{\infty}\frac{\featindex^{-2\alpha}}{1+nz^{-1}\featindex^{-\alpha})^2}}{1-\frac{n}{z^2}\sum\limits_{\featindex=1}^{\infty}\frac{\featindex^{-2\alpha}}{(1+nz^{-1}\featindex^{-\alpha})^2}}.
\end{equation}
Therefore, the excess prediction error suggestively decomposes into two terms, the first accounting for the variance due to sampling, while the second reflects the additional variance entailed by the label noise. Unlike a typical bias-variance decomposition, the effect of the bias (as manifested by the $\lambda$-dependent $z$ term) is subsumed in both terms. For simplicity, the first term in the numerator shall be referred to in the rest of the derivation as the \textit{sample variance term}, and the second sum in the numerator as the \textit{noise variance term}.

In the same limit, the equation defining $z$ \eqref{eq:app:z} is amenable to being rewritten:
\begin{equation}
\label{eq:app:z2}
  z=n\lambda +\frac{z}{\samples }\sum\limits_{\featindex=1}^\infty \frac{1}{1+\frac{z}{\samples }\featindex^\alpha} ,
\end{equation}
or, approximating the Riemann sum by an integral
\begin{align}
\label{eq:app:z3}
    z&\approx \samples\lambda+\left(\frac{z}{\samples }\right)^{1-\frac{1}{\alpha}}\int_{\left(\frac{z}{\samples }\right)^{1/\alpha}}^\infty \frac{\dd x}{1+x^\alpha}.
\end{align}

\subsection{Infinite sample limit and the scaling of the generalisation error}
\label{subsection:app:nbig}
Consider now the limit $n\gg 1$ with $\lambda$ scaling with n
\begin{equation}
    \lambda\sim n^{-\regdecay}.
\end{equation}
Note that the scalings of $z$ with respect to $\samples$ differ according to the regularisation $\lambda$, depending on which of the two terms on the right hand side of equation \eqref{eq:app:z2} dominates. If the first $\samples\lambda$ term dominates, then \eqref{eq:app:z2} simplifies to $z\approx\samples\lambda$. For this to be self-consistent, we must have $(z/n)^{1-\frac{1}{\alpha}}\approx\lambda^{1-\frac{1}{\alpha}}\ll\samples\lambda$, i.e. $n\gg \lambda^{-\frac{1}{\alpha}}$. In the converse case where the second term in \eqref{eq:app:z2} dominates, $z\sim \samples ^{1-\alpha}$. For this to consistently hold, one needs $(z/n)^{1-\frac{1}{\alpha}}\approx n^{1-\alpha}\gg\samples\lambda$, i.e. $n\ll \lambda^{-\frac{1}{\alpha-1}}$.
Depending on which term dominates in \eqref{eq:app:z2}, two regime may be distinguished:
\begin{itemize}
    \item In the \textit{effectively non-regularized} $\regdecay>\alpha$ regime, $n\ll \lambda^{-\frac{1}{\alpha}}$ so $z\sim \samples ^{1-\alpha}$. In this regime the regularization totally disappears from the analysis and KRR behaves just as if $\lambda=0$.
    \item in the \textit{effectively regularized} $\regdecay<\alpha$ regime, $n\gg \lambda^{-\frac{1}{\alpha}}$ regime, $z\approx \samples\lambda$.
\end{itemize}

\subsection{Effectively non-regularized regime}
\label{subsection:app:unreg}
\paragraph{Sample variance term: }  As before, depending on $1+2r\alpha,\alpha$, it is sometimes possible to rewrite the sample variance term in integral form. If $r<1$, \begin{align}
   \sum\limits_{\featindex=1}^{\infty}
    \frac{\featindex^{-1-2r\alpha}}{(1+\samples z^{-1}\featindex^{-\alpha})^2}&\sim
    \samples ^{-2r\alpha}\sum\limits_{\featindex=1}^{\infty}
    \frac{\left(\frac{\featindex}{\samples }\right)^{-1-2r\alpha}}{(1+\left(\frac{\featindex}{n}\right)^{-\alpha})^2}\frac{1}{\samples }
    \sim \samples ^{-2r\alpha}\int\limits_0^\infty \frac{x^{-1+2(1-r)\alpha}}{(1+x^\alpha)^2}=\mathcal{O}(\samples ^{-2r\alpha}).
\end{align}
If $r>1$, it is no longer possible to write the Riemann sum as an integral, and 
\begin{align}
   \sum\limits_{\featindex=1}^\infty
    \frac{\featindex^{-1-2r\alpha}}{(1+\samples z^{-1}\featindex^{-\alpha})^2}&=
    \sum\limits_{\featindex=1}^\samples 
    \frac{\featindex^{-1-2r\alpha}}{(1+\samples ^\alpha \featindex^{-\alpha})^2}
    +
    \samples ^{-2r\alpha}\sum\limits_{\featindex=\samples }^\infty
    \frac{\left(\frac{\featindex}{\samples }\right)^{-1-2r\alpha}}{(1+\left(\frac{\featindex}{n}\right)^{-\alpha})^2}\frac{1}{\samples }=\mathcal{O}(\samples ^{-2\alpha}).
\end{align}
\paragraph{Noise variance term: } It is possible to similarly decompose the sum in the noise variance term to find
\begin{equation}
    \frac{\samples \sigma^2}{z^2}\sum\limits_{\featindex=1}^{\infty}
    \frac{\featindex^{-2\alpha}}{(1+\samples z^{-1}\featindex^{-\alpha})^2}=\mathcal{O}(\sigma^2).
\end{equation}
From this, it follows that:
\begin{itemize}
    \item for $n\ll\sigma^{-\frac{2}{2\alpha\mathrm{min}(r,1)}}$ the sample variance term dominates the numerator, and
\begin{equation}
\label{eq:app:scaling1}
    \epsilon_g-\sigma^2=\mathcal{O}(\samples ^{-2\alpha\mathrm{min}(r,1)})
\end{equation}
\item for $n\gg\sigma^{-\frac{2}{2\alpha\mathrm{min}(r,1)}}$ the noise variance term dominates the numerator, and determines the decay of the excess prediction error
\begin{equation}
\label{eq:app:scaling11}
    \epsilon_g-\sigma^2=\mathcal{O}(\sigma^2)
\end{equation}
\end{itemize}
These two subregimes are amenable to being written in the more compact form \eqref{eq:green_red_decay}:
\begin{equation}
    \epsilon_g-\sigma^2=\mathcal{O}\left(\mathrm{max}\left(\sigma^2,\samples ^{-2\alpha\mathrm{min}(r,1)}\right)\right)
\end{equation}

\subsection{Effectively regularized regime}
\label{subsection:app:reg}
\paragraph{Sample variance term: } By the same token, in the second $\regdecay<\alpha$ regularized regime, provided $r<1$, one can write the sample variance term as a Riemann sum (since $\lambda\sim n^{-\regdecay}=o(1)$):

\begin{align}
   \sum\limits_{\featindex=1}^{\infty}
    \frac{\featindex^{-1-2r\alpha}}{(1+\samples z^{-1}\featindex^{-\alpha})^2}&\sim
    \lambda^{2r}
    \sum\limits_{\featindex=1}^{\infty}\frac{(\featindex\lambda^{\frac{1}{\alpha}})^{-1+2(1-r)\alpha}}{\left((\featindex\lambda^{\frac{1}{\alpha}})^\alpha+1\right)^2}\lambda^{\frac{1}{\alpha}}\sim \lambda^{2r}\int\limits_0^\infty \frac{x^{-1+2(1-r)\alpha}}{(1+x^\alpha)^2}\nonumber\\
    &=\mathcal{O}(\samples ^{-2\regdecay r}).
\end{align}
In the $r>1$ case, 
\begin{align}
   \sum\limits_{\featindex=1}^\infty
    \frac{\featindex^{-1-2r\alpha}}{(1+\samples z^{-1}\featindex^{-\alpha})^2}&=
    \sum\limits_{\featindex=1}^{\samples ^\frac{\regdecay}{\alpha}}
    \frac{\featindex^{-1-2r\alpha}}{(1+\frac{1}{\lambda}\featindex^{-\alpha})^2}
    +
    \lambda^{\frac{-2r\alpha}{\alpha}}
    \sum\limits_{\featindex=\samples ^\frac{\regdecay}{\alpha}}^\infty\frac{(\featindex\lambda^{\frac{1}{\alpha}})^{-1+2(1-r)\alpha}}{\left((\featindex\lambda^{\frac{1}{\alpha}})^\alpha+1\right)^2}\lambda^{\frac{1}{\alpha}}.
\end{align}
Upper and lower bounds can be straightfowardly found for the first sum and the following equivalence established
\begin{equation}
    \sum\limits_{\featindex=1}^{\samples ^\frac{\regdecay}{\alpha}}
    \frac{\featindex^{-1-2r\alpha}}{(1+\frac{1 }{\lambda}\featindex^{-\alpha})^2}
    \sim \samples ^{-2\regdecay}\sum\limits_{\featindex=1}^{\samples ^\frac{\regdecay}{\alpha}}\featindex^{-1+2(1-r)\alpha}=\mathcal{O}(\samples ^{-2\regdecay}),
\end{equation}
while the second sum is a Riemann sum of order $\mathcal{O}(\samples ^\frac{(-2r\alpha)\regdecay}{\alpha})=o(\samples ^{-2\regdecay})$. Therefore, the first sum in the numerator scales like 
\begin{equation}
    \sum\limits_{\featindex=1}^\infty
    \frac{\featindex^{-1-2r\alpha}}{(1+\samples z^{-1}\featindex^{-\alpha})^2}=\mathcal{O}\left(\samples ^{-2\regdecay\rm{min}(r,1)}\right)
\end{equation}
\paragraph{Noise variance term: } The scaling of the noise variance term is found along similar lines to be 
\begin{equation}
    \frac{\samples\sigma^2 }{z^2}\sum\limits_{\featindex=1}^{\infty}
    \frac{\featindex^{-2\alpha}}{(1+\samples z^{-1}\featindex^{-\alpha})^2}=\mathcal{O}(\sigma^2\samples ^\frac{\regdecay-\alpha}{\alpha}).
\end{equation}
\\
\noindent If the noise variance term decays faster in $\samples$, then the sample variance term always dominates (since $\sigma^2$ is at most $\mathcal{O}(1)$). This is the case when 
\begin{equation}
    0<\regdecay<\frac{\alpha}{2\alpha\mathrm{min}(r,1)+1}
\end{equation}
and then the generalization excess prediction error scales like
\begin{equation}
    \epsilon_g-\sigma^2=\mathcal{O}\left(\samples ^{-2\regdecay\rm{min}(r,1)}\right).
\end{equation}
In the case where $\alpha>\regdecay>\frac{\alpha}{2\alpha\mathrm{min}(r,1)+1}$ there exist two regimes depending on how $\samples$ compares with the noise strength
\begin{itemize}
    \item if $n\ll\sigma^{\frac{2}{1-\frac{\regdecay}{\alpha}(1+2\alpha\mathrm{min}(r,1))}}$ the sample variance term dominates and we recover the noiseless case
    \begin{equation}
    \epsilon_g-\sigma^2=\mathcal{O}\left(\samples ^{-2\regdecay\rm{min}(r,1)}\right).
\end{equation}
\item if $n\gg\sigma^{\frac{2}{1-\frac{\regdecay}{\alpha}(1+2\alpha\mathrm{min}(r,1))}}$ the noise variance term dominates and 
    \begin{equation}
    \epsilon_g-\sigma^2=\mathcal{O}(\sigma^2\samples ^\frac{\regdecay-\alpha}{\alpha}).
\end{equation}
\end{itemize}

All those regimes can be written more compactly as \eqref{eq:blueorange_decay} 
\begin{equation}
\label{eq:app:scaling2}
    \epsilon_g-\sigma^2=\mathcal{O}\left(\mathrm{max}\left(\sigma^2,n^{1-2\regdecay\rm{min}(r,1)-\frac{\regdecay}{\alpha}}\right)
    \samples ^\frac{\regdecay-\alpha}{\alpha}
    \right).
\end{equation}

\paragraph{Case $\regdecay<0$: } We give here for completeness the case in which the regularization grows with $n$. Then the sample variance term scales like
\begin{equation}
    \sum\limits_{\featindex=1}^{\infty}
    \frac{\featindex^{-1-2r\alpha}}{(1+\samples z^{-1}\featindex^{-\alpha})^2}=\mathcal{O}(1).
\end{equation}
To see this, use a lower and upper bound starting from $0\le nz^{-1}\featindex^{-\alpha}\sim n^{\regdecay}\featindex^{-\alpha}\le 1 $ for all $\featindex\ge 1$ and all $n$. The noise variance term scales like
\begin{equation}
    \frac{\sigma^2\samples }{z^2}\sum\limits_{\featindex=1}^{\infty}
    \frac{\featindex^{-2\alpha}}{(1+\samples z^{-1}\featindex^{-\alpha})^2}\sim \sigma^2 n^{2\regdecay-1}=o(1),
\end{equation}
meaning 
\begin{equation}
    \epsilon_g-\sigma^2=\mathcal{O}(1)
\end{equation}

\subsection{Continuity across the regularization crossover line}

The $\regdecay=\alpha$ is actually comprised in the $\regdecay>0$ case of the $\regdecay<\alpha$ regimes. On the $\regdecay=\alpha$ separation line, there is no discontinuity between the non-regularized exponents and the regularized exponents, since
\begin{equation}
    \mathrm{max}\left(\sigma^2,n^{1-2\regdecay\rm{min}(r,1)-\frac{\regdecay}{\alpha}}\right)
    \samples ^\frac{\regdecay-\alpha}{\alpha}
    \overset{\regdecay=\alpha}{=}
    \mathrm{max}\left(\sigma^2,\samples ^{-2\alpha\mathrm{min}(1,r)}\right).
\end{equation}

\subsection{Asymptotically optimal regularization}

The derivation in subsections \ref{subsection:app:unreg} and \ref{subsection:app:reg} effectively delimit the four regimes in Fig.~\ref{fig:phase_diagram}: the effectively non-regularized noiseless green regime, the effectively regularized noiseless blue regime, the effectively non-regularized noisy red regime, and the effectively regularized noisy orange regime.

For any given $n$, we define the \textit{asymptotically optimal} $\regdecay$ as the regularization decay yielding fastest decay of the excess prediction error. This corresponds to finding the $\regdecay$ with maximal excess error decay along a vertical line at abscissa $\samples$ in the phase diagram Fig.~\ref{fig:phase_diagram}.

If $n\gg n_1^\star\approx\sigma^{-\frac{1}{\alpha\mathrm{min}(r,1)}}$ (effectively noisy regime), the noise-induced crossover line is crossed for 
\begin{equation}
    \regdecay_c\approx\left(1-2\frac{\ln \sigma}{\ln n}
    \right)\frac{\alpha}{1+2\alpha\mathrm{min}(r,1)}.
\end{equation}
The asymptotically optimal $\regdecay^\star$ is found as
\begin{align}
    \regdecay^\star=\underset{\regdecay}{\mathrm{argmax}}\left(
    2\regdecay\mathrm{min}(r,1)\mathbbm{1}_{0<\regdecay<\regdecay_c}+\frac{\alpha-\regdecay}{\alpha}\mathbbm{1}_{\regdecay_c<\regdecay<\alpha}+0\times \mathbbm{1}_{\alpha<\regdecay}
    \right).
\end{align}
Since the argument of the argmax is an increasing function of $\regdecay$ on $(0,\regdecay_c)$ and a decreasing function on $(\regdecay_c,\infty)$ the maximum is found for $\regdecay^\star=\regdecay_c$. The corresponding decay for the excess error is 
\begin{equation}
    \mathrm{max}\left(2\regdecay^\star\mathrm{min}(r,1),\frac{\alpha-\regdecay^\star}{\alpha}\right)=2\regdecay^\star\mathrm{min}(r,1)\approx\frac{2\alpha\mathrm{min}(r,1)}{1+2\alpha\mathrm{min}(r,1)}\left(1-2\frac{\ln \sigma}{\ln n}\right).
\label{eq:app:illdef}
\end{equation}
 It is nonetheless ill-defined to talk about an aymptotically optimal rate for the regularization that continuously varies with $\samples$ when $\samples$ is comparable with $\sigma^{-2}$, since \eqref{eq:app:illdef} means that the excess error is not even a power law in this region. An asymptotic statement can be however made. For $\samples\gg n_2^\star\approx\mathrm{max}(n_1^\star,\sigma^{-2})$,
\begin{equation}
    \regdecay^\star\approx\frac{\alpha}{1+2\alpha\mathrm{min}(r,1)},
\end{equation}
and the excess error decays like \eqref{eq:opt_noisy}
\begin{equation}
    \epsilon_g^\star-\sigma^2=\mathcal{O}\left(n^{-\frac{2\alpha\mathrm{min}(r,1)}{1+2\alpha\mathrm{min}(r,1)}}\right).
\end{equation}

For $n\ll\sigma^{-\frac{2}{2\alpha\mathrm{min}(r,1)}}$ (effectively noiseless regime), we have
\begin{align}
    \regdecay^\star=\underset{\regdecay}{\mathrm{argmax}}\left(
    2\regdecay\mathrm{min}(r,1)\mathbbm{1}_{0<\regdecay<\alpha}+2\alpha\mathrm{min}(r,1)\mathbbm{1}_{\alpha<\regdecay}
    \right),
\end{align}
which means that any $\regdecay^\star\in (\alpha, \infty)$ is optimal (in particular, vanishing regularization is optimal), and we recover \eqref{eq:opt_noiseless}
\begin{equation}
    \epsilon_g^\star-\sigma^2=\mathcal{O}\left(n^{-2\alpha\mathrm{min}(r,1)}\right).
\end{equation}

\section{A dictionary of notation in the literature}
\label{appendix:dic}
While the capacity and source conditions are assumed in almost all works concerned with the decay rates of Kernel methods, the actual notations for the capacity and source terms $\alpha, r$ greatly vary. We provide in this appendix a table summarizing notations for the references \cite{bordelon2020, Berthier2020TightNC,pillaud2018statistical,spigler2019asymptotic,Caponnetto2005FastRF,caponnetto2007optimal,steinwart2009optimal,fischer2017sobolev,Jun2019KernelTR}

\begin{table}[h]
\centering
\begin{tabular}{ |p{2cm}||p{2cm}|p{2cm}|  }
 \hline
Reference & $\alpha$\cite{pillaud2018statistical}&$r$\cite{pillaud2018statistical}\\
 \hline
\cite{bordelon2020}&$b$&$\frac{a-1}{2b}$\\
\hline
\cite{spigler2019asymptotic}&$\frac{\alpha_S}{d}$&$\frac{1}{2}(\frac{\alpha_T}{\alpha_S}-d)$\\
 \hline
\cite{Berthier2020TightNC}& $\beta$&$\frac{2\delta+\beta-1}{2\beta}$\\
\hline
\cite{Caponnetto2005FastRF,caponnetto2007optimal}&$b$&$\frac{c}{2}$\\
\hline
\cite{steinwart2009optimal,fischer2017sobolev}&$\frac{1}{p}$&$\frac{\beta}{2}$\\
\hline
\cite{Jun2019KernelTR}&$b$&$\beta$\\
\hline
\end{tabular}
\caption{Dictionary between different notations previously used in the KRR literature.}
\label{table:dic}
\end{table}

\section{Details on real data sets}
\label{appendix:real_data}

\subsection{Feature map to diagonal covariance for real datasets}

In the general case where the data $x$ is drawn from a generic distribution $\rho_x$, we remind the equations defining the feature map $\feature$ \eqref{eq:def_feature_map}:

\begin{align}
    &\feature{}(x)=\Sigma^{\frac{1}{2}}\phi(x)\\
    & \mathbb{E}_{x\sim \rho_x} \left[\phi(x)\phi(x)^T\right]=\mathbbm{1}_\featuredim{}\\
    & \mathbb{E}_{x'\sim \rho_x} \left[K(x,x')\phi(x')\right]=\Sigma \phi(x)
\end{align}

In the of a real dataset $\mathcal{D}=\{x^\mu,y^\mu\}_{\mu=1}^{\samples_\tot}$ from which both the train and test set are uniformly drawn, the distribution is then the empirical uniform distribution over $\mathcal{D}$,
\begin{equation}
    \rho_x(\cdot)=\frac{1}{\samples_\tot}\sum\limits_{\mu=1}^{\samples_\tot}\delta(\cdot-x^\mu).
\end{equation}
Defining the Gram matrix $(K_{\mu\nu})_{\mu,\nu=1}^{\samples_\tot}\overset{\mathrm{def}}{=}(K(x^\mu,x^\nu))_{\mu,\nu=1}^{\samples_\tot}\in\mathbb{R}^{\samples_\tot\times \samples_\tot}$, the equations defining the feature map \eqref{eq:def_feature_map} can be rewritten in the simpler matricial form

\begin{align}
    \feature=\phi\Sigma^{\frac{1}{2}}, &&
    \frac{1}{\samples_\tot}\phi^T\phi=\mathbbm{1}_{\samples_\tot}, &&
    \frac{1}{\samples_\tot}K\phi= \phi\Sigma
\end{align}
\noindent where $\phi,\feature,\lambda,K \in\mathbb{R}^{\samples_\tot\times \samples_\tot} $, and the feature space is of dimension $p=\samples_\tot$, with the $\mu^{\mathrm{th}}$ line of $\feature{}$ (resp. $\phi$) corresponding to $\feature{}(x^\mu)$ (resp. $\phi(x^\mu)$). To access the coordinates $\theta^\star_\featindex$ in the basis of the features $\feature$, remember $\feature\theta^\star=y$, hence
\begin{equation}
    \theta^\star=\frac{1}{\samples_\tot}\Sigma^{-1}\feature^Ty
\end{equation}

\subsection{Estimation of source and capacity}
\begin{figure}
    \centering
    \includegraphics[scale=0.44]{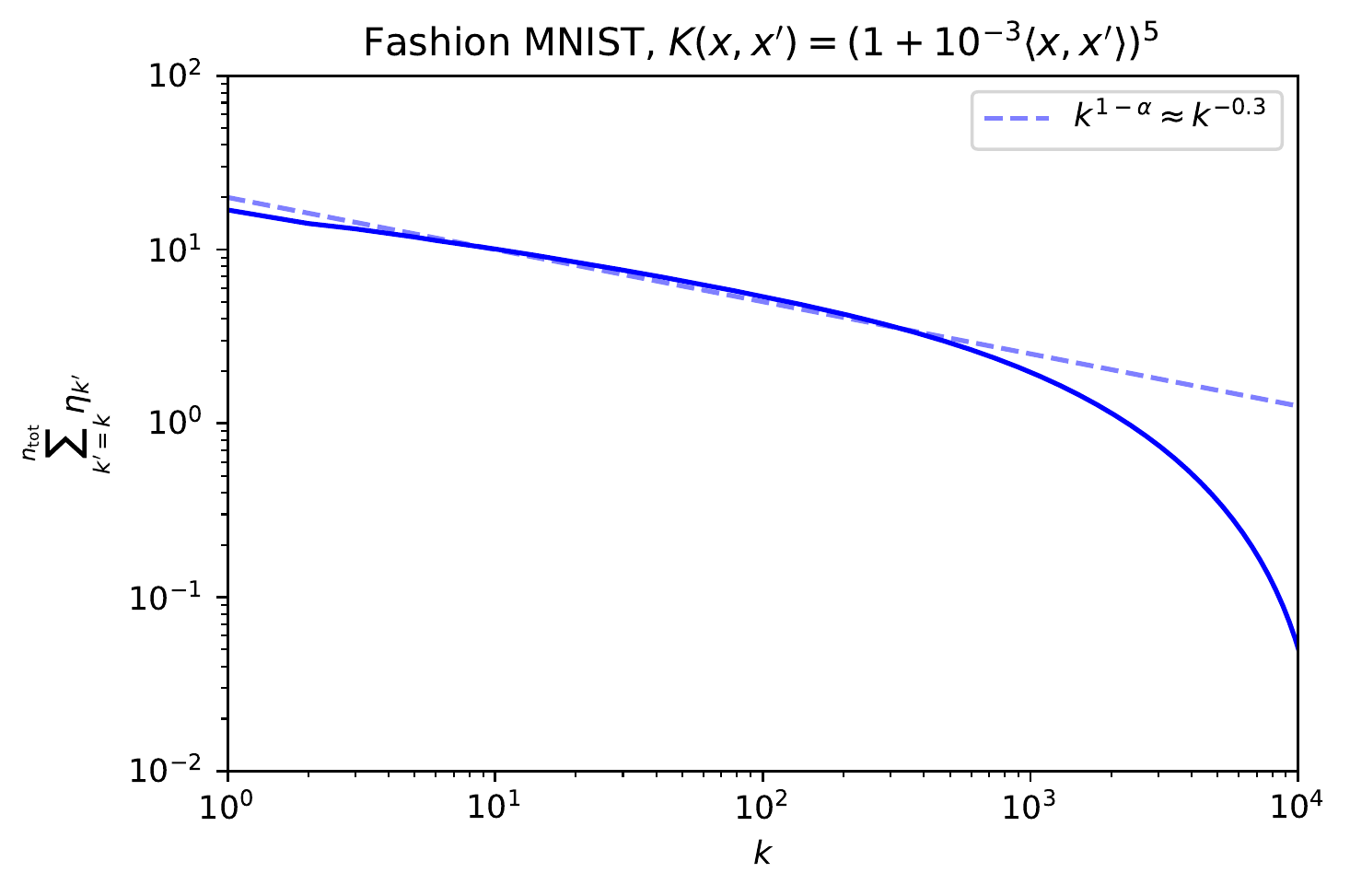}
    \includegraphics[scale=0.44]{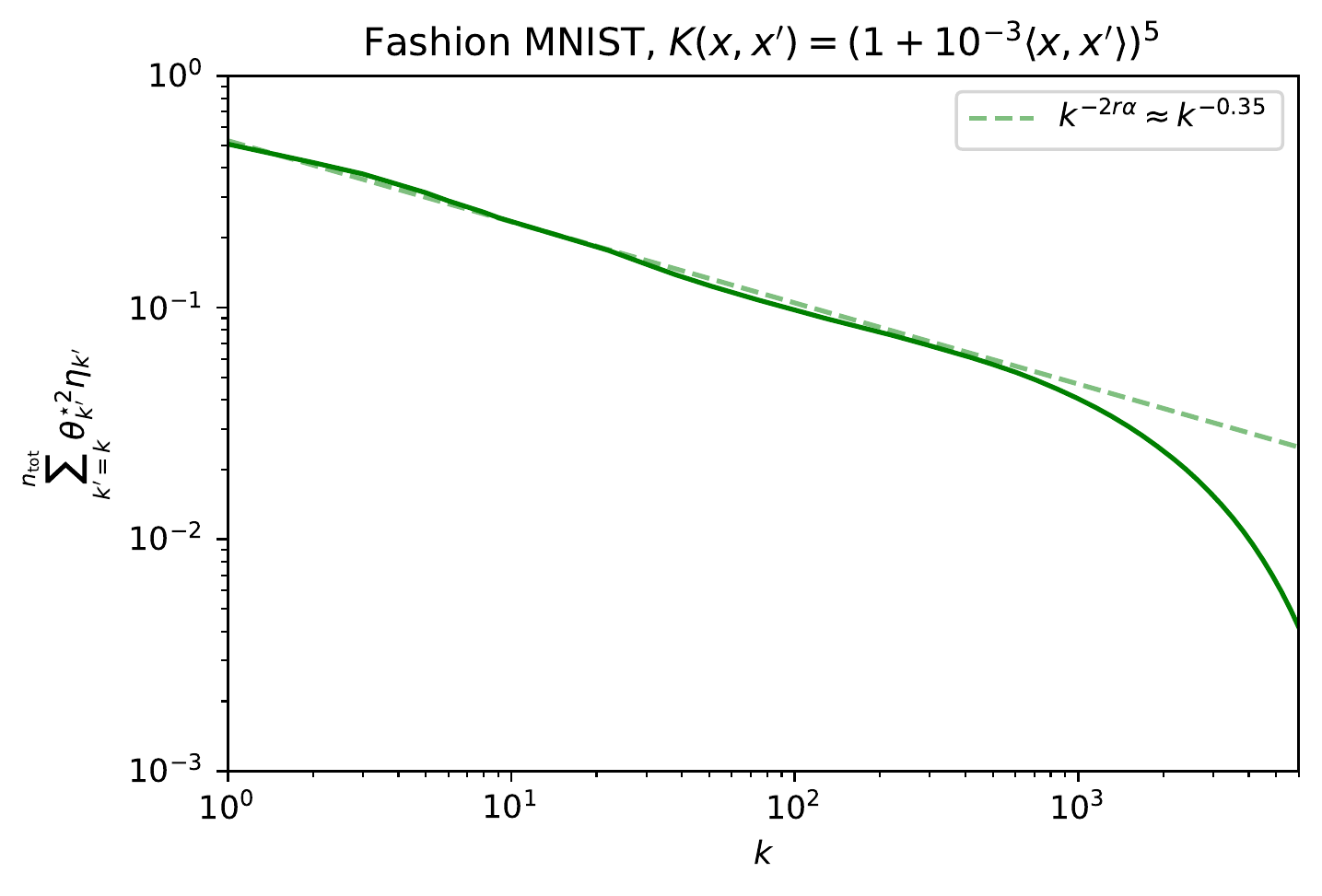}
    \includegraphics[scale=0.44]{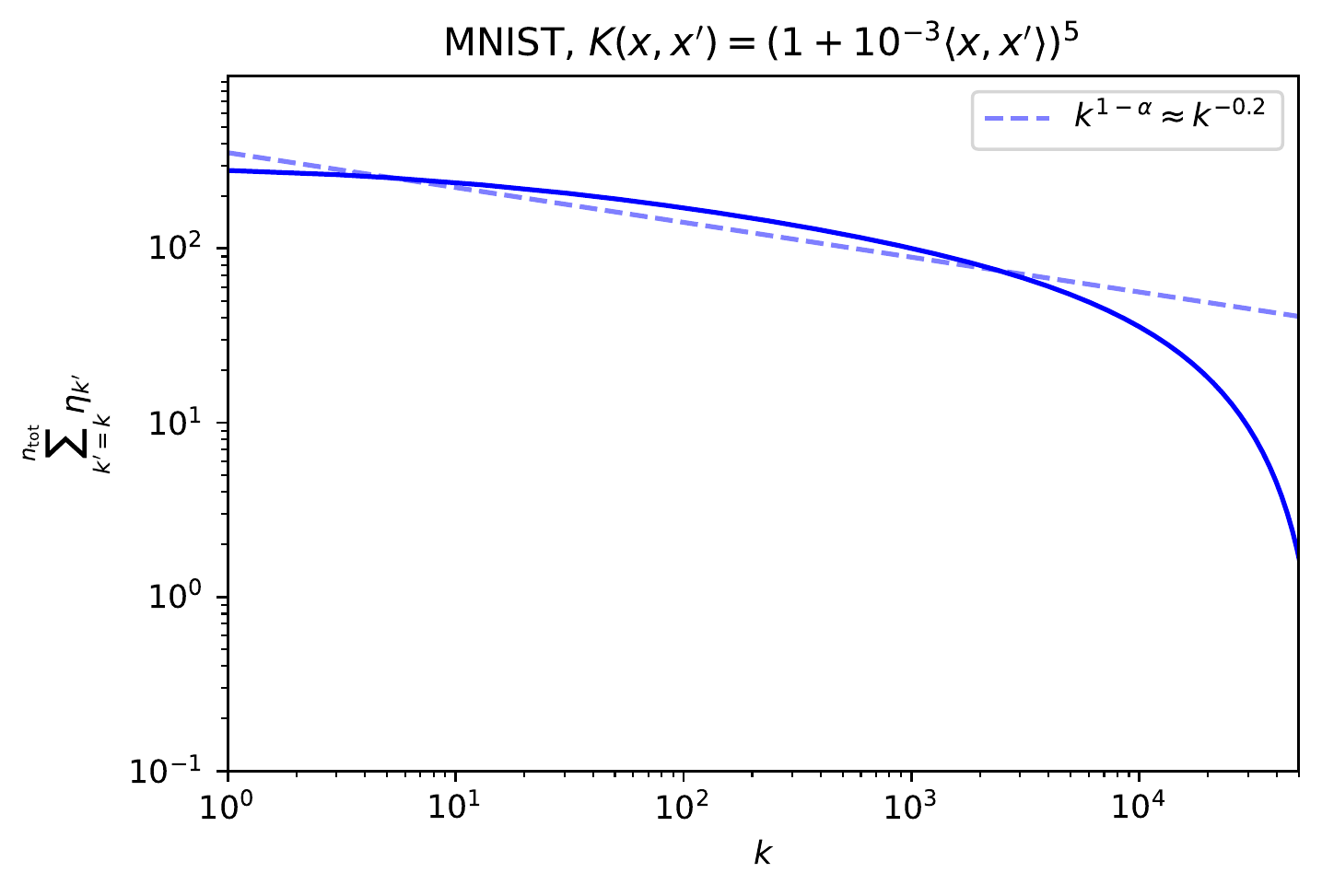}
    \includegraphics[scale=0.44]{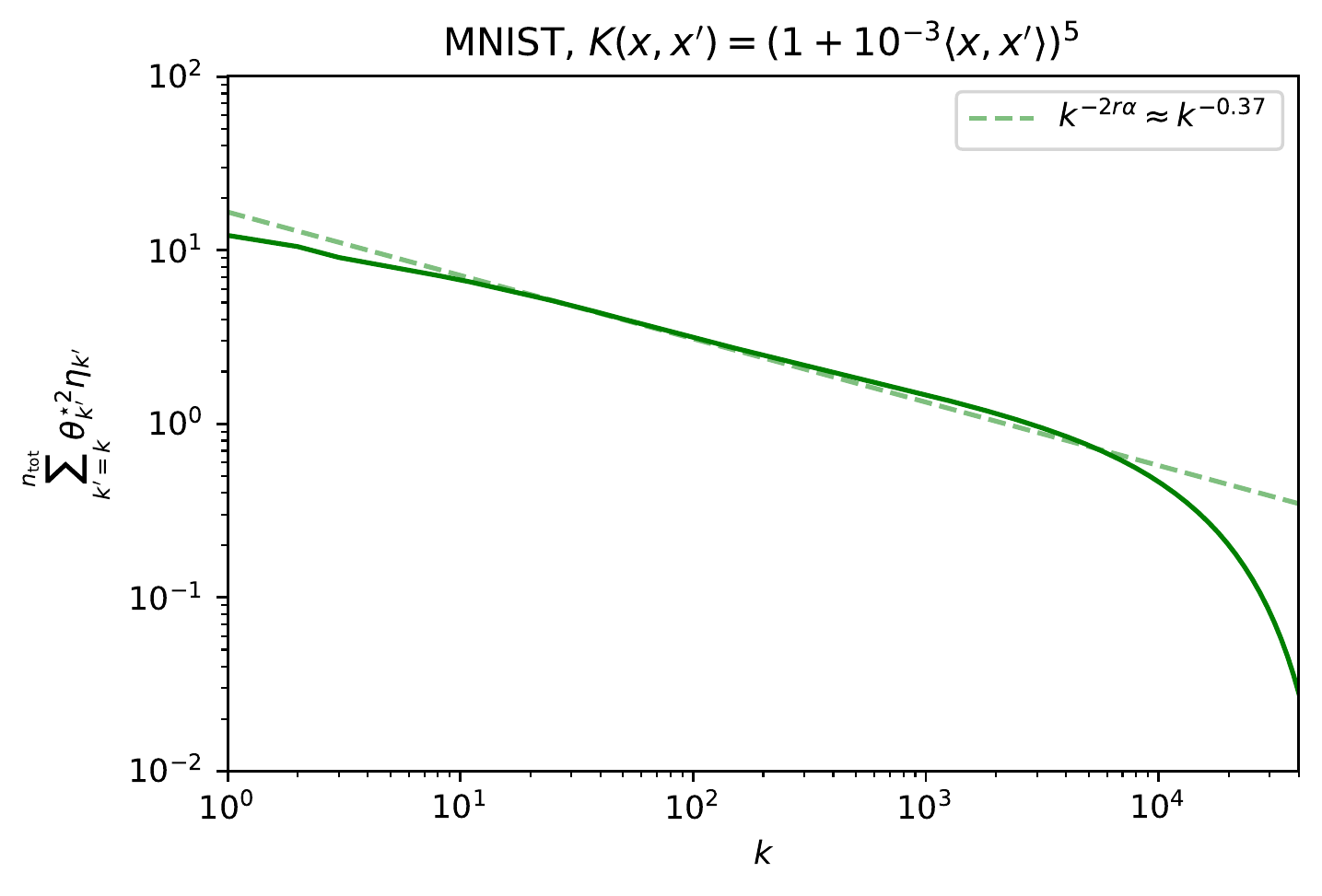}
    \includegraphics[scale=0.44]{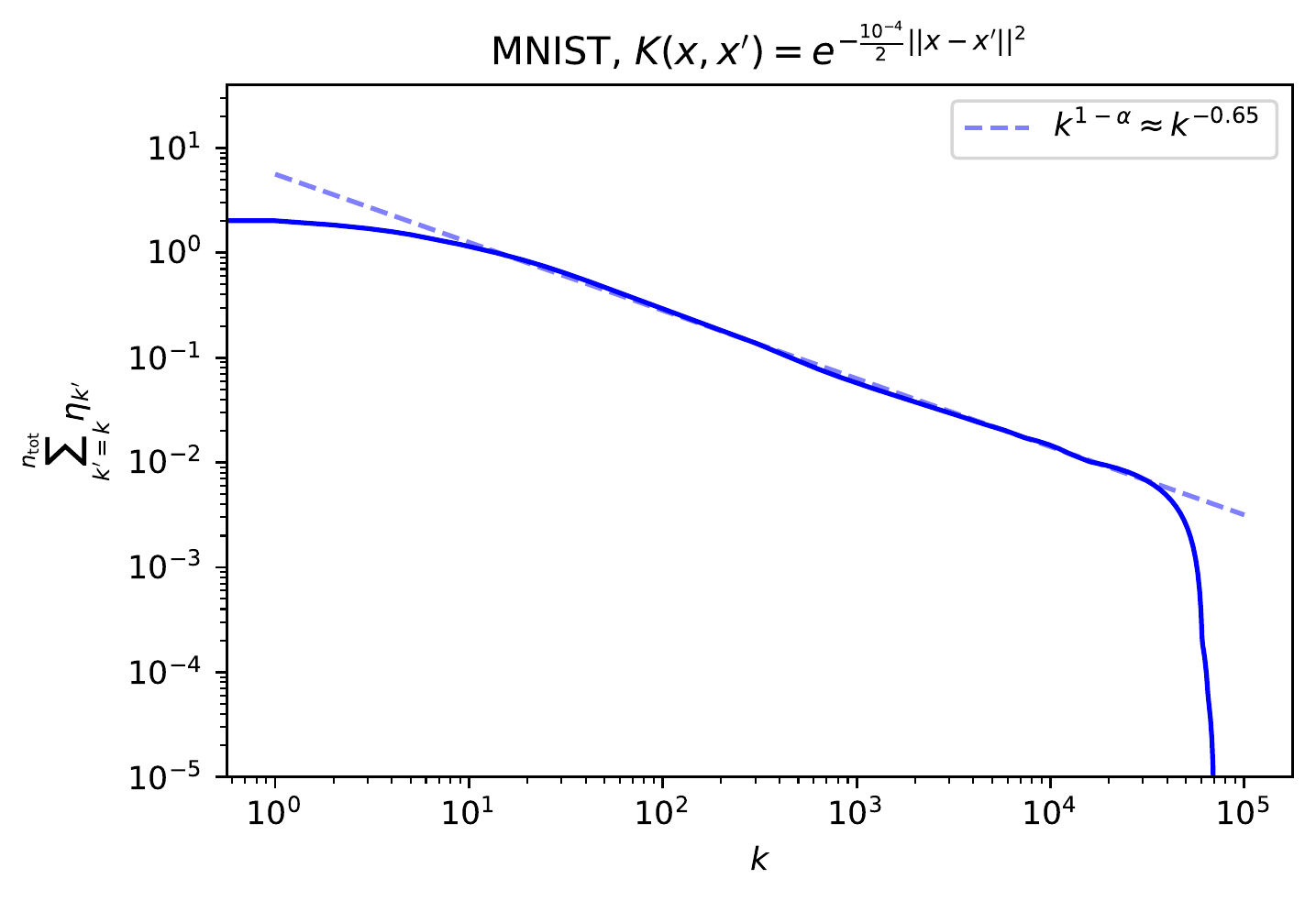}
    \includegraphics[scale=0.44]{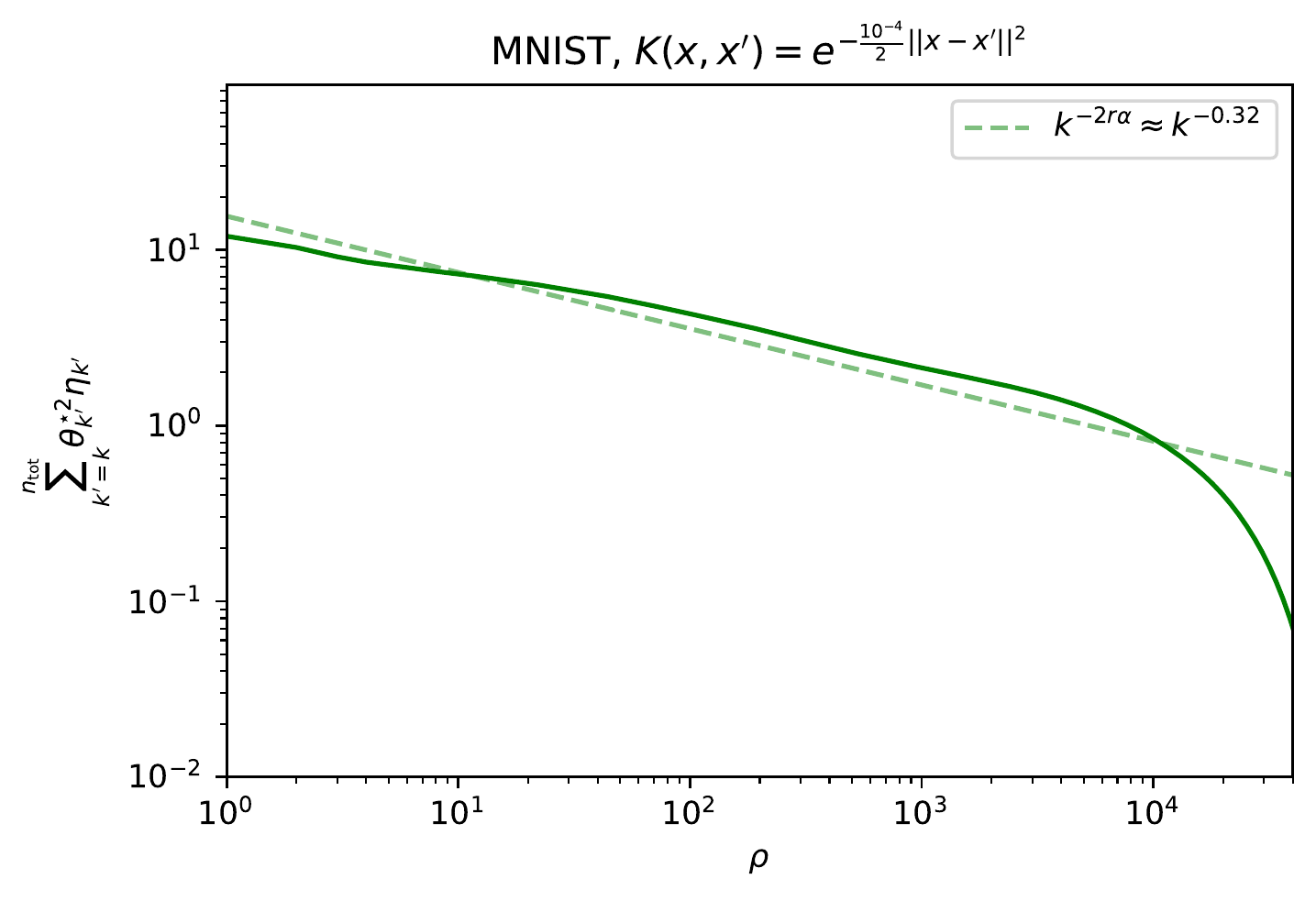}
    \includegraphics[scale=0.44]{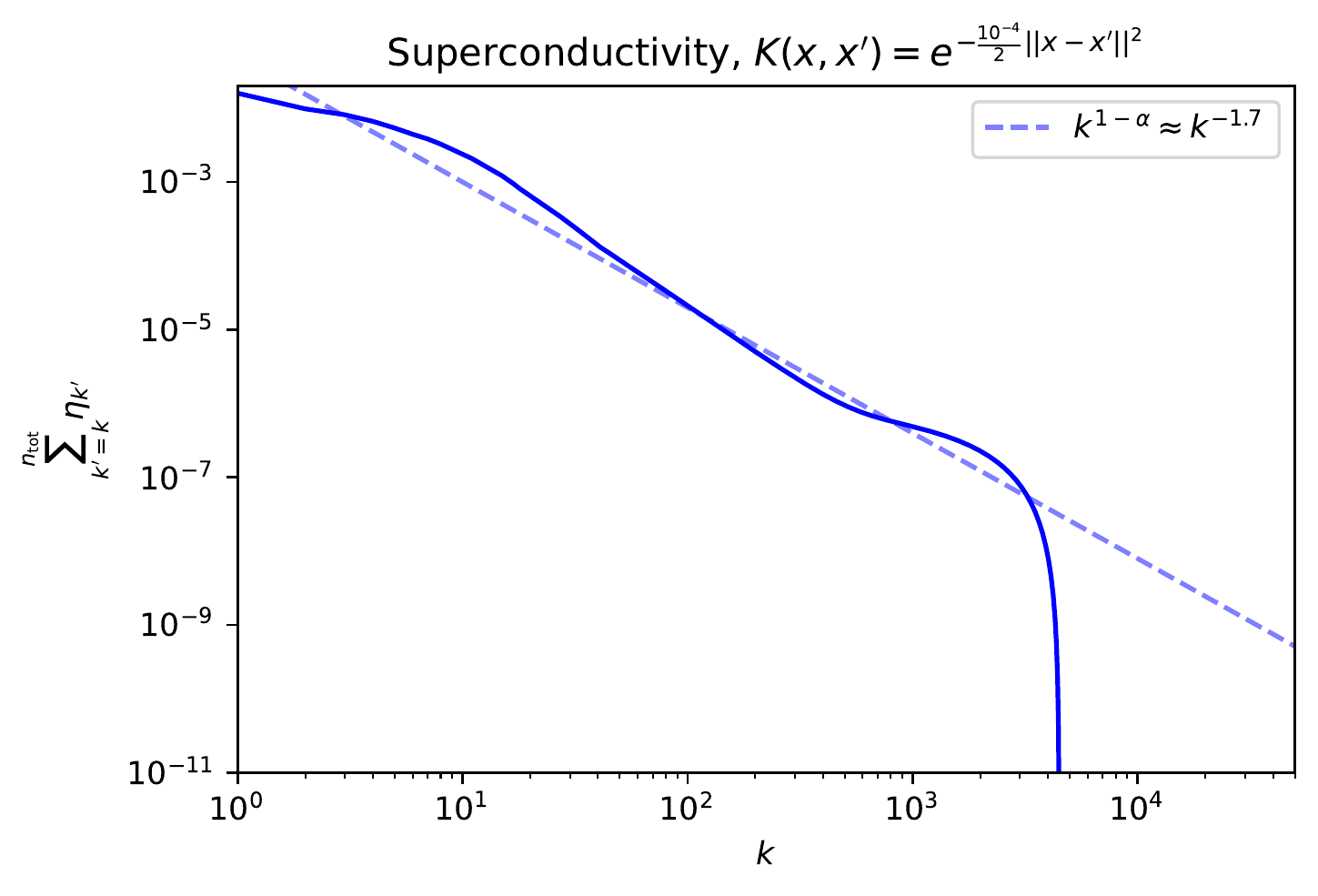}
    \includegraphics[scale=0.44]{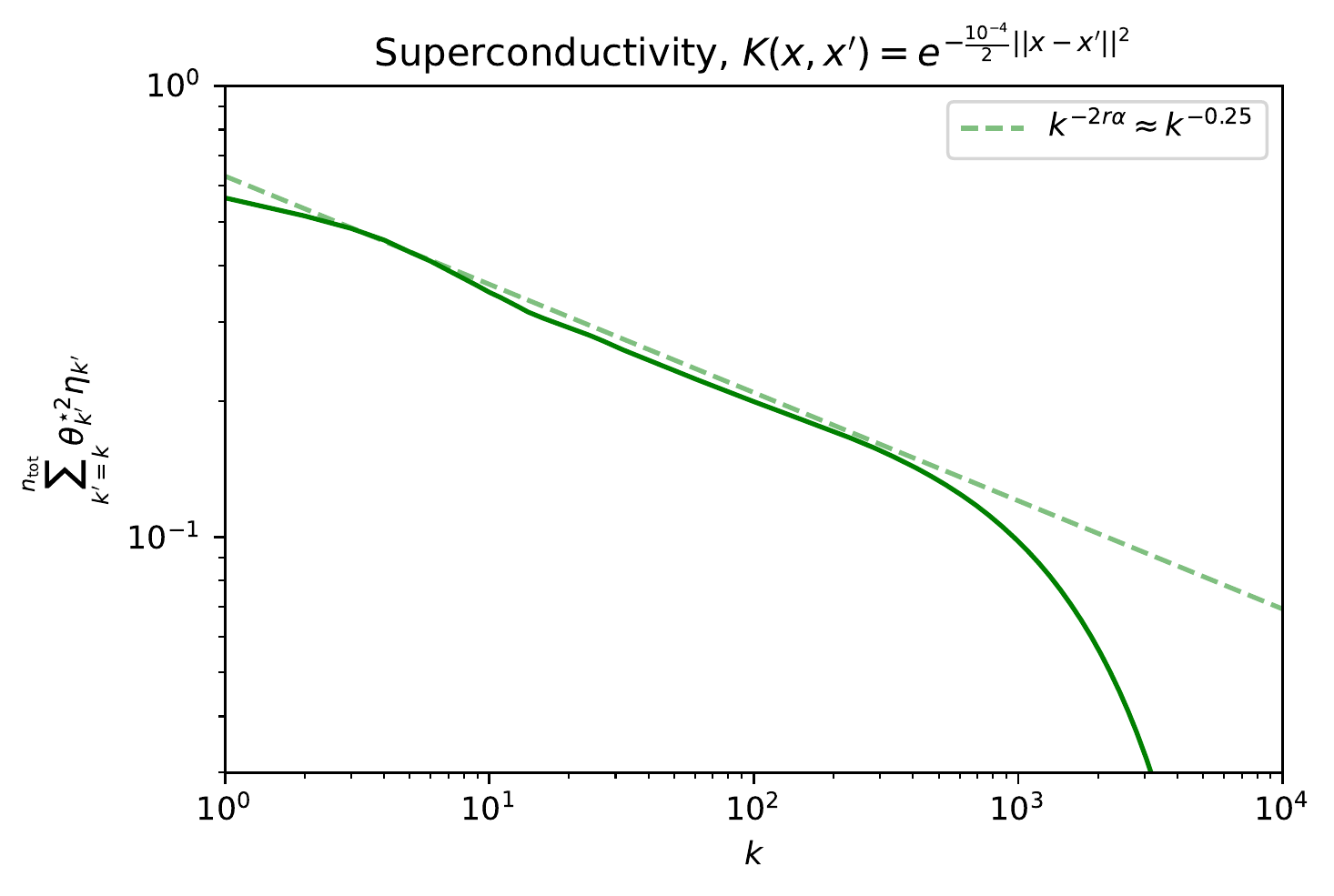}
    \caption{Measurement of empirical values for capacity and source $\alpha, r$ for real datasets (Fashion MNIST t-shirt and coat, MNIST) and RBF, polynomial kernels. Because the feature space is of finite dimension $\samples_\tot$ all the curves exhibit a sharp drop at $n_\tot$. A power-law was fitted on the functions \eqref{eq:app:cumu} on a range of $\featindex$ where these looked reasonnably like power-laws.}
    \label{fig:Measure_params}
\end{figure}

The capacity and source terms $\alpha, r $ can be empirically estimated for the dataset $\mathcal{D}$ from the eigenvalues $\{\lambda_\featindex\}_{\featindex=1}^{\samples_\tot}$ of the Gram matrix $K$ and the components $\{\theta^\star_\featindex\}_{\featindex=1}^{\samples_\tot}$ of the teacher vector. Supposing decays like \eqref{eq:model_def}, the cumulative functions read:
\begin{align}
    \sum\limits_{\featindex'=\featindex}^{\samples_\tot}\lambda_{\featindex'}\sim \featindex^{1-\alpha},
    &&
    \sum\limits_{\featindex'=\featindex}^{\samples_\tot}\lambda_{\featindex'}\theta^{\star 2}_{\featindex'}\sim \featindex^{-2r\alpha}.
\label{eq:app:cumu}
\end{align}
These functions are plotted in Fig.~\ref{fig:Measure_params} and the terms $\alpha, r $ estimated therefrom. The use of the cumulative functions, rather than a direct estimation from the coordinates, allows the integration to smoothen out the curves and get a more consistent estimation. The values of $\alpha, r $ thereby measured are summarized in Table.~\ref{table:measuredalphar}. Note that the power-law form \eqref{eq:model_def} and the assumption $p=\infty$ fail to hold for real data, and the series \eqref{eq:app:cumu} have power-law form only on a range of indices $\featindex$, before a sharp drop due to the finite dimensionality $\samples_\tot$ of the feature space, see Fig.~\ref{fig:Measure_params}. The range of indices $\featindex$ where the power-law form \eqref{eq:model_def} seems to hold was qualitatively assessed, and linear regression run thereon to estimate $\alpha, r$. Since there is no clear objective way to determine the range the fit should be conducted on, the estimates slightly vary depending on the precise choice of the regression range, without however overly hurting the qualitative agreement with simulations Fig.~\ref{fig:Real_CV} and \ref{fig:Real_noreg}.

\subsection{Details on simulations}
We close this appendix by providing further detail on the simulations on real data (Figs.~\ref{fig:Real_CV} and \ref{fig:Real_noreg}).

For each simulation at sample size $\samples$ a train set was created by subsampling $\samples$ samples from the total available dataset $\mathcal{D}$. To mitigate the effect of spurious correlations due to sampling a finite dataset, the whole dataset $\mathcal{D}$ has been used as a test set, following \cite{Loureiro2021CapturingTL, Canatar2021SpectralBA}. A kernel ridge regressor was fitted on the train set with the help of the \texttt{scikit-learn KernelRidge} package \cite{scikit-learn}. For Fig.~\ref{fig:Real_CV}, the best regularization $\reg$ was estimated using the \texttt{scikit-learn GridSearchCV} \cite{scikit-learn} default $5-$fold cross-validation procedure on the Grid $\reg\in\{0\}\cup (10^{-10},10^{5})$, with logarithmic step size $\delta\mathrm{log}\reg=0.026$. The excess test error was averaged over $10$ independent samplings of the train set and noise realizations.

\section{More crossovers}
\label{appendix:Crossover}
\begin{figure}
    \centering
    \includegraphics[scale=0.7]{Plots/Phase_diagram_prefactor.pdf}
    \caption{Phase diagram for $\lambda_0\ll 1$ and $\sigma<\lambda_0^{\mathrm{min}(r,1)+\frac{1}{\alpha}}$.As for Fig.~\ref{fig:phase_diagram}. The solid red line corresponds to the noise-type crossover line, while the blue line indicates the regularization-type crossover line. Note that for low enough values of the regularization, the two crossover lines can be intercepted by a same horizontal line. This means that a double crossover is in theory observable (green-blue-orange), the first induce by regularization (see also \cite{bordelon2020}) and the second being noise-induced.}
    \label{fig:diagram_prefactor_1}
\end{figure}

\begin{figure}
    \centering
    \includegraphics[scale=0.7]{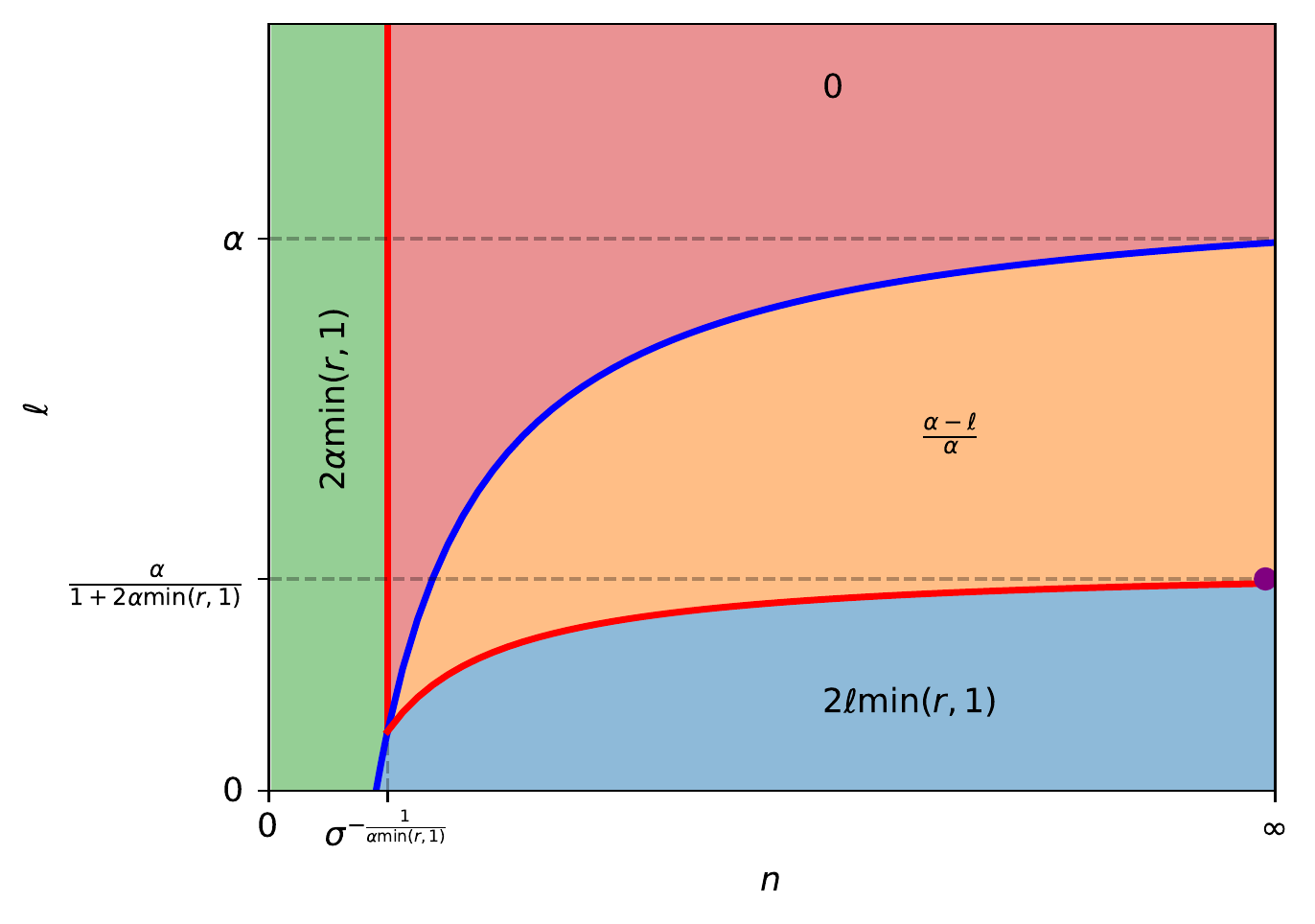}
    \caption{Phase diagram for $\lambda_0\ll 1$ and $\sigma>\lambda_0^{\mathrm{min}(r,1)+\frac{1}{\alpha}}$. As for Fig.~\ref{fig:phase_diagram}, the asymptotically optimal decays $\regdecay^\star$ are indicated in solid red. The solid red line corresponds to the noise-type crossover line, while the blue line indicates the regularization-type crossover line. Note that for low enough values of the regularization, the blue crossover line can be intercepted by a horizontal line, alongside the red crossover line twice. Consequently a triple crossover is in theory observable (green-red-orange-blue), with two noise-induced and one regularization-induced.}
    \label{fig:diagram_prefactor_2}
\end{figure}

\subsection{Regularization-induced crossover}
On top of the distinction between effectively noiseless regimes (green, blue regions in Fig.~\ref{fig:phase_diagram}) and effectively noisy regimes (red, orange in Fig.~\ref{fig:phase_diagram}), the four regimes can also be classified in effectively non-regularized (green, red) and effectively regularized (blue, orange), see also the discussion in Section \ref{sec:derivation}. In Fig.~\ref{fig:phase_diagram}, the non-regularized regions lie above the horizontal separation line $\regdecay=\alpha$, while the regularized ones lie below. In this appendix, we discuss a more generic setting for which this separation line ceases to be horizontal, thereby creating \textit{a new crossover line}. Similarly to the noise-type crossover line discussed in the main text, a learning curve that crosses this regularization-induced crossover line transitions from an non-regularized regime (green, red) to a regularized one (blue, orange), characterized by differing decays for the excess error $\epsilon_g-\sigma^2$. In Fig.~\ref{fig:diagram_prefactor_1} and Fig.~\ref{fig:diagram_prefactor_2}, the noise-type crossover line is depicted in red, while the regularization-type crossover line is in blue.

In this section we thus detail the more general setting 
\begin{align}
    \lambda=\lambda_0 \samples^{-\regdecay},
 \label{eq:lambda_prefactor}
\end{align}
with, compared to the setup studies in the main text and Appendix~\ref{appendix:computations}, an additional prefactor $\lambda_0$ to the regularization $\lambda$ that is allowed to be $\ll1$. The particular case $\regdecay=0, \lambda_0$ small, has been studied in \cite{bordelon2020}, and has been shown to give rise to a crossover due to the regularization, on top on the one evidenced in the present work due to the noise. 
\begin{itemize}
    \item For small $\samples$, KRR focuses on fitting the spiked subspace comprising large variance dimensions, and satisfies the norm constraint introduced by the regularization on the lower importance subspace. This phenomenon can be loosely regarded as the bias version of the benign overfitting for noise variance (\cite{Bartlett2020BenignOI, Tsigler2020BenignOI}) : the bias induced by the loss of expressivity due to the norm constraint is effectively diluted over less important dimensions, thereby not impacting the generalization. 
    \item For larger $\samples$, decreasing the excess error $\epsilon_g-\sigma^2$ requires a good KRR fit also on the subspace of lesser importance, and the regularization effect is felt. In a noiseless green-blue crossover, this results in a slower decay because of the bias introduced by regularizing. On a noisy red-orange crossover, the regularization conversely helps to mitigate the noise and enables the excess risk to decay again.
\end{itemize}

\subsection{Outline of the computation}
The derivation for the general case \eqref{eq:lambda_prefactor} follows very closely Appendix~\ref{appendix:computations}.

\begin{itemize}
    \item If $n\ll\lambda_0^{-\frac{1}{\alpha-\regdecay}}$ or $\regdecay>\alpha$, $n\ll \lambda^{-\frac{1}{\alpha}}$ so $z\sim \samples ^{1-\alpha}$,
    \item If $n\gg\lambda_0^{-\frac{1}{\alpha-\regdecay}}$ and $\regdecay<\alpha$, $n\gg \lambda^{-\frac{1}{\alpha}}$ and $z\approx \lambda$.
    \item If $n=\lambda_0^{-\frac{1}{\alpha-\regdecay}}$ regime and $\regdecay<\alpha$, $n\sim \lambda^{-\frac{1}{\alpha}}$ so $z\sim \lambda\sim \samples ^{1-\alpha}\sim \samples ^{-\regdecay}$,
\end{itemize}
Note that the introduction of $\lambda_0\ll 1$ means the limits between regularized and non-regularized regime are now involving both $\samples, \regdecay$ as opposed to just $\regdecay$ in Appendix ~\ref{appendix:computations} (see also Fig.~\ref{fig:phase_diagram}). In the first $n\ll\lambda_0^{-\frac{1}{\alpha-\regdecay}}$ regime, the regularization effect is not sensed and the computation is identical to the $\lambda_0=1$ case in Appendix~\ref{appendix:computations}. In the regularized $n\gg\lambda_0^{-\frac{1}{\alpha-\regdecay}}$, keeping track of the prefactors yields
\begin{align}
    \epsilon_g-\sigma^2=\mathcal{O}\left(\lambda_0^{2\mathrm{min}(r,1)}n^{-2\regdecay\mathrm{min}(r,1)}\right)+\mathcal{O}\left(\sigma^2n^{\frac{\regdecay-\alpha}{\alpha}}\lambda_0^{\frac{-1}{\alpha}}\right),
\end{align}
so the excess risk decays like
\begin{equation}
\epsilon_g-\sigma^2=\mathcal{O}
\left( \lambda_0^{2\mathrm{min}(r,1)}n^{\frac{\regdecay-\alpha}{\alpha}}\mathrm{max}\left(\left(\frac{\sigma}{\lambda_0^{\mathrm{min}(r,1)+\frac{1}{\alpha}}}
\right)^2,n^{-2\regdecay\mathrm{min}(r,1)+1-\frac{\regdecay}{\alpha}}
\right)
\right).
\label{eq:prefactor_smallreg_excessrisk}
\end{equation}
Depending on whether the maximum in \eqref{eq:prefactor_smallreg_excessrisk} is always realized by one of its two arguments, or by one then the other as $\samples$ is varied, there may be a noise-induced crossover.
\begin{itemize}
    \item if $\sigma<\lambda_0^{\mathrm{min}(r,1)+\frac{1}{\alpha}}$ and $\regdecay\le \frac{\alpha}{1+2\alpha\mathrm{min}(r,1)}$, the second argument of the maximum in \eqref{eq:prefactor_smallreg_excessrisk} dominates for all $n\ge 1$ so no crossover is to be observed (see Fig.~\ref{fig:diagram_prefactor_1}), and
    \begin{equation}
        \epsilon_g-\sigma^2=\mathcal{O}\left(\lambda_0^{2\mathrm{min}(r,1)}n^{-2\regdecay\mathrm{min}(r,1)}\right).
        \label{eq:prefactor_smalldecay_decay1}
    \end{equation}
    \item if $\sigma>\lambda_0^{\mathrm{min}(r,1)+\frac{1}{\alpha}}$ and $\regdecay\ge \frac{\alpha}{1+2\alpha\mathrm{min}(r,1)}$, the first argument of the maximum in \eqref{eq:prefactor_smallreg_excessrisk} dominates for all $n\le 1$ so no crossover is to be observed (see Fig.~\ref{fig:diagram_prefactor_2}), and
    \begin{equation}
        \epsilon_g-\sigma^2=\mathcal{O}\left(\sigma^2n^{\frac{\regdecay-\alpha}{\alpha}}\lambda_0^{\frac{-1}{\alpha}}\right).
    \label{eq:prefactor_smalldecay_decay2}
    \end{equation}
    \item in any other case, a crossover between the decays \eqref{eq:prefactor_smalldecay_decay1} and \eqref{eq:prefactor_smalldecay_decay2} is observed, at a sample size
    \begin{equation}
        \samples^\star_1=\left(\frac{\sigma}{\lambda_0^{\mathrm{min}(r,1)+\frac{1}{\alpha}}}
\right)^{\frac{2}{1-\frac{\regdecay}{\alpha}(1+2\alpha\mathrm{min}(r,1))}}.
    \end{equation}
    The crossover is from \eqref{eq:prefactor_smalldecay_decay1} to  \eqref{eq:prefactor_smalldecay_decay2} if $\regdecay\ge \frac{\alpha}{1+2\alpha\mathrm{min}(r,1)}$ an in the other order if $\regdecay\le \frac{\alpha}{1+2\alpha\mathrm{min}(r,1)}$.
\end{itemize}

The determination of the asymptotically optimal decays carries through as Appendix~\ref{appendix:computations}, with the same conclusions. The four regimes and their respective limit, as well as the optimal $\regdecay$ at very large $\samples$ (purple point), are summarized in Figs.~\ref{fig:diagram_prefactor_1} and \ref{fig:diagram_prefactor_2}.

\subsection{Double and triple crossovers}
\begin{figure}
    \centering
    \includegraphics[scale=0.6]{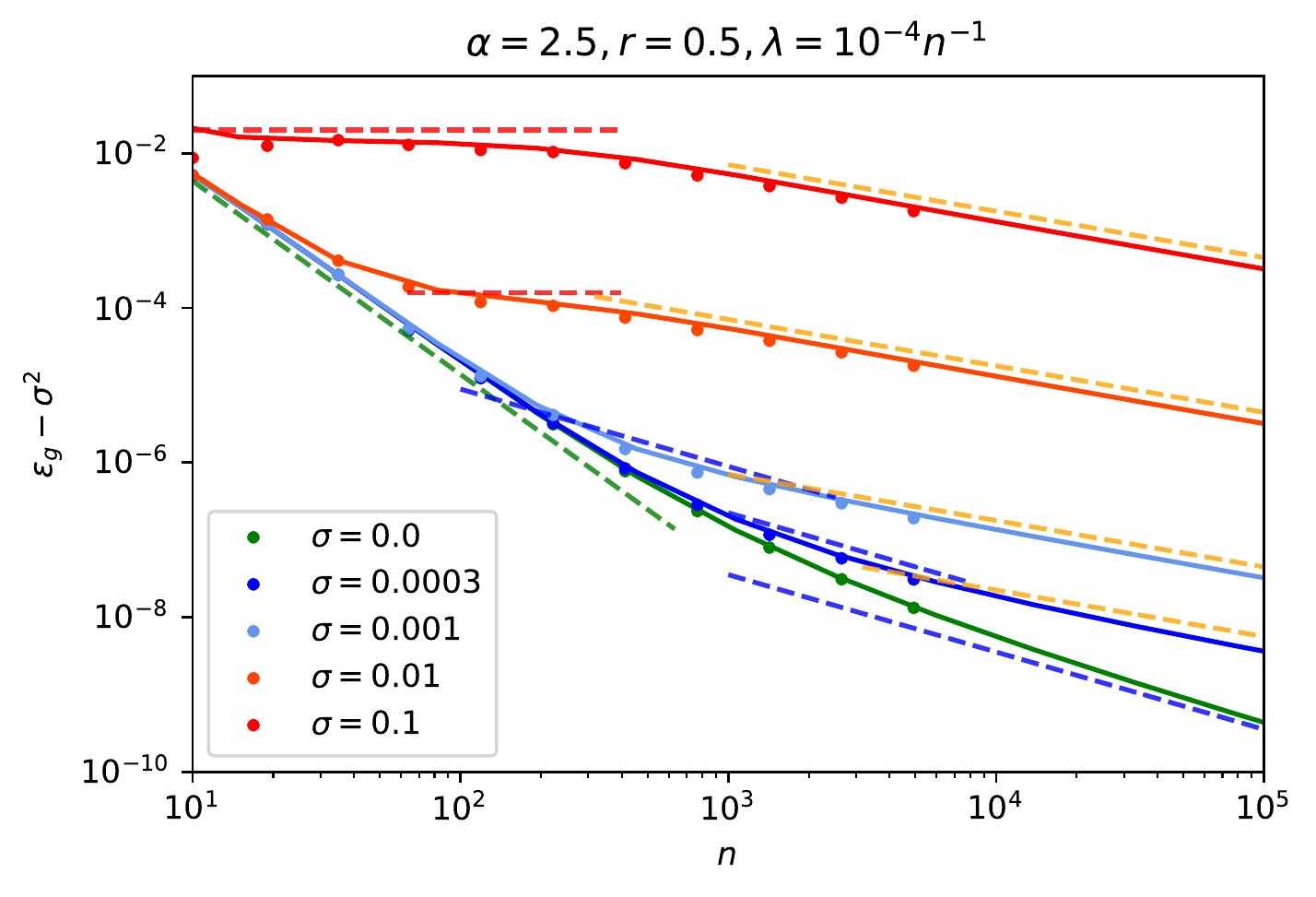}
    \includegraphics[scale=0.5]{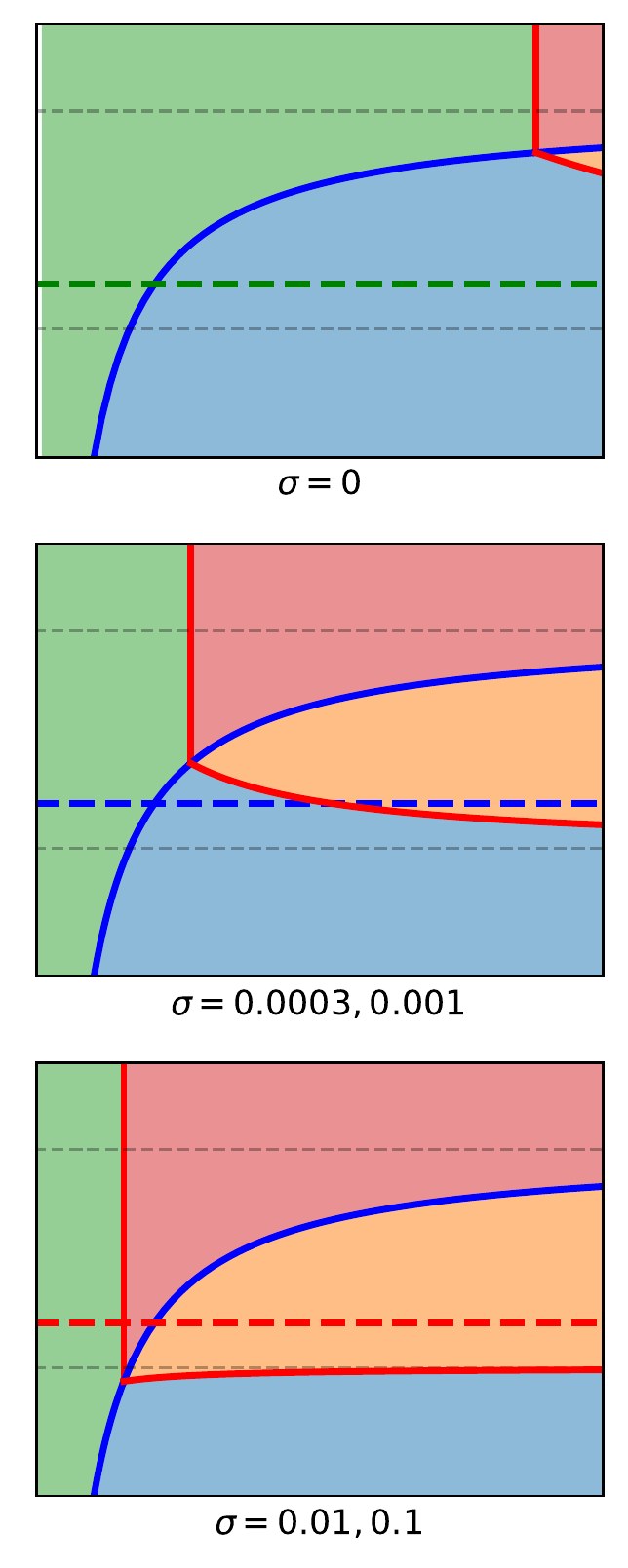}
    \caption{Excess risk learning curves for $\alpha=2.5,~r=0.5,~\lambda_0=10^{-4},~\regdecay=1$. The noise level $\sigma$ is varied and the corresponding phase diagrams given on the right. For $\sigma=0$ (green curve and top diagram on the right), a simple regularization-induced crossover (green-to-blue) is observed. For $\sigma=3.10^{-4}$ and $\sigma=10^{-3}$ (blue curves on the left, middle diagram on the right) a double crossover green-to-blue-to-orange is observed. The first is regularization induced, while the second is due to noise. For $\sigma=10^{-2}$ and $\sigma=10^{-1}$ (orange, red curves and bottom diagram), a double green-to-red-to-orange crossover is observed, respectively noise and regularization induced.}
    \label{fig:triple}
\end{figure}
We therefore recover the regularization induced crossover reported in \cite{bordelon2020} for the special case $\regdecay=0,\sigma=0$. It corresponds to the green-to-blue transition for the lowest $\regdecay$ in Fig.~\ref{fig:diagram_prefactor_1}~\ref{fig:diagram_prefactor_2}. We stress that such a mechanism is entirely due to the regularization, and hence happens \textit{on top} of the noise-induced crossover studied in the present work. It is therefore possible in theory to observe both crossovers in succession.

We detail as an example a double green-to-blue-to-orange crossover (see blue curves in Fig.\ref{fig:triple}). For small $\samples$ (non-regularized noiseless green regime), KRR fits the heavy dimensions. Both noise overfitting and bias are benign. As $\samples$ is increased, the blue regularization type crossover line in Fig.~\ref{fig:diagram_prefactor_1} is crossed and the regularized noiseless blue region entered. More of the less important dimensions need to be fitted well: bias is felt and entails a slower decay, but the noise overfitting is diluted over even less important dimensions and remains benign. As the red noise-type crossover line is passed into the regularized noisy orange region, the overfitting ceases to be benign and hurts the decay rate.

\section{Relation to worst-case bounds}
\label{appendix:related}
In this section, we sketch informally how the blue and orange exponents \eqref{eq:blueorange_decay} can also be derived from the worst case bounds \cite{Lin2018OptimalRF,Jun2019KernelTR}. Note that the recovery from worst case bounds of the exponents \eqref{eq:blueorange_decay}, which were here derived in the gaussian design setting, suggests that for these regimes the worst case exponents are also equal to the typical case exponents. We remind the reader that this has also already been shown to be the case for the asymptotically optimal lambda, see section \ref{section:phase_diagram}, exponents \eqref{eq:opt_noisy} and \cite{Caponnetto2005FastRF,caponnetto2007optimal}.

\subsection{ Optimal rates for spectral algorithms with least-squares regression over Hilbert spaces \cite{Lin2018OptimalRF}}

To relate the notations employed in \cite{Lin2018OptimalRF} to ours, the correspondances
\begin{align}
    \gamma\in]0,1]=\frac{1}{\alpha} && \zeta\in[0,1]=r && \theta\in[0,1]=1-\ell
\end{align}
\begin{align}
    \mathcal{L}=\Sigma && f_H=f^\star
\end{align}
should be used, see also section \ref{appendix:dic}. With respect to their equation (18) defining the source condition, the setting considered in the present work corresponds to the special case $\phi(u)=u^\zeta$. Note that the assumption $\ell\le 1$ is slightly more restrictive than those employed in this paper.
The main result of \cite{Lin2018OptimalRF} is their Theorem 4.2, which in our notations translates to
\begin{theorem}
(Theorem 4.2 1) \cite{Lin2018OptimalRF}) With probability $1-\delta$, there exist constants $\Tilde{C}_1,\Tilde{C}_2,\Tilde{C}_3$ st
\begin{equation}
    (\epsilon_g-\sigma^2)^{\frac{1}{2}}=||f^\star-\hat{f}||_{\mathcal{H}}=||\theta^\star-\hat{w}||_{\mathcal{L}^2}\le\left(
    \Tilde{C}_1 n^{-\mathrm{max}(\frac{1}{2},1-r)}+\Tilde{C}_2n^{-\frac{1}{2}}\lambda^\frac{-1}{2\alpha}+\Tilde{C}_3\lambda^r
    \right)\ln\frac{6}{\delta}\left(\ln\frac{6}{\delta}+\frac{\mathrm{max}(\frac{1}{1-l},\ln n)}{\alpha}
    \right)
\end{equation}
viz. expliciting the scalings
\begin{equation}
    (\epsilon_g-\sigma^2)^{\frac{1}{2}}\le\left(
    \Tilde{C}_1 n^{-1+\mathrm{min}(\frac{1}{2},\mathrm{min}(r,1))}+\Tilde{C}_2n^{-\frac{1}{2}\frac{\alpha-\ell}{\alpha}}+\Tilde{C}_3n^{-\ell\mathrm{min(r,1)}}
    \right)\ln\frac{6}{\delta}\left(\ln\frac{6}{\delta}+\frac{\mathrm{max}(\frac{1}{1-l},\ln n)}{\alpha}
    \right)
\label{eq:theorem_Lin}
\end{equation}
\end{theorem} 
we replaced $\zeta$ by $\mathrm{min}(r,1)$ since \cite{Lin2018OptimalRF} work under the assumption $\zeta=r\in[0,1]$ in order to make contact with the exponents in the present paper. Up to logarithmic corrections, one recognizes the blue ($\Tilde{C}_3$ term in \eqref{eq:theorem_Lin}) and orange ($\Tilde{C}_2$ term in \eqref{eq:theorem_Lin}) exponents, the effectively unregularized red and green regimes \eqref{eq:green_red_decay} being inaccessible in this setting because of the restriction $\ell\le1$. One can further show that the first $\Tilde{C}_1$ term in \eqref{eq:theorem_Lin} is always subdominant, since:
\begin{itemize}
    \item if $r\ge\frac{1}{2}$, $n^{-\frac{1}{2}+\frac{\ell}{2\alpha}}\gg n^{-\frac{1}{2}}$ and the $\Tilde{C}_2$ term dominates the $\Tilde{C}_1$ term
    \item if $r\le\frac{1}{2}$, $n^{-1+\mathrm{min}(1,r)}\ll n^{-\frac{1}{2}}\ll n^{-\ell\mathrm{min}(r,1)} $ since $\ell r\le r\le \frac{1}{2}$ and the $\Tilde{C}_3$ term dominates $\Tilde{C}_1$ term
\end{itemize}
The relative competition between the $\Tilde{C}_{2,3}$ contributions in \eqref{eq:theorem_Lin} determine the bleu to orange crossover. This suggests in particular that typical and worst case coincide within these regimes.

\subsection{Kernel Truncated Randomized Ridge Regression:
Optimal Rates and Low Noise Acceleration \cite{Jun2019KernelTR}}

Similar to \cite{Lin2018OptimalRF}, the notations translate to 
\begin{align}
    \beta\in [0,\frac{1}{2}]=r&&b\in [0,1]=\frac{1}{\alpha}.
\end{align}
Note that in \cite{Jun2019KernelTR}, it is further assumed that the labels are bounded by a constant $Y$ while this only holds with high probability in our setting.
The Theorem $3$ in \cite{Jun2019KernelTR} then reads:
\begin{theorem} (Informal)
The error gap given by the KTR$^3$ algorithm \cite{Jun2019KernelTR} is approximately bounded by, for any $\epsilon_r,\epsilon_\alpha>0$, for the power law ansatz \eqref{eq:app:decayansatz}:
\begin{equation}
    \epsilon_g-\sigma^2\le \lambda^{2r-2\epsilon_r}\frac{1}{2\alpha\epsilon_r}+\mathrm{min}\left[\frac{4Y^2}{\alpha\epsilon_\alpha\lambda^{\frac{1}{\alpha}+\epsilon_\alpha}n}
    \mathrm{min}\left(
    \mathrm{ln}\left(1+\frac{1}{\lambda}\right)^{1-\frac{1}{\alpha}-\epsilon_\alpha},\frac{\alpha}{1+\alpha\epsilon_\alpha}
    \right),
    \frac{\lambda^{2r-2\epsilon_r-1}}{2\alpha\epsilon_r n}+\frac{\sigma^2}{\lambda n}
    \right],
\end{equation}
so in the particular setting $\lambda=n^{-\ell}$
\begin{equation}
    \epsilon_g-\sigma^2\le n^{-2r\ell+2\ell\epsilon_r}\frac{1}{2\alpha\epsilon_r}+\mathrm{min}\left[\frac{4Y^2}{\alpha\epsilon_\alpha}n^{-\frac{\alpha-\ell}{\alpha}+\epsilon_\alpha\ell}
    \mathrm{min}\left(
    \mathrm{ln}\left(1+n^\ell\right)^{1-\frac{1}{\alpha}-\epsilon_\alpha},\frac{\alpha}{1+\alpha\epsilon_\alpha}
    \right),
    \frac{n^{-2r\ell+2\ell\epsilon_r+\ell -1}}{2\alpha\epsilon_r }+\sigma^2 n^{-1+\ell}
    \right].
\end{equation}
\end{theorem}
\noindent If $\sigma\ne 0$, the $\sigma^2 n^{-1+\ell}$ term dominates in the second argument of the min and the min is realized by its first argument, leading to
\begin{equation}
    \epsilon_g-\sigma^2=\mathcal{O}(n^{-2\ell r})+\mathcal{O}(n^{-\frac{\alpha-\ell}{\alpha}})
\end{equation}
namely the blue/orange crossover \eqref{eq:blueorange_decay}. If $\sigma=0$ the bound is necessarily looser than $\mathcal{O}(n^{-2\ell r})$ which is coherent since in the noiseless setting only the blue exponent can be observed.

\end{document}